\documentclass[runningheads]{llncs}

\usepackage{eccv}
\usepackage{eccvabbrv}
\usepackage{graphicx}
\usepackage{booktabs}
\usepackage{multirow}
\usepackage[accsupp]{axessibility}
\usepackage{hyperref}
\usepackage{orcidlink}
\usepackage{marvosym}

\begin{document}

\title{Fabric Image Demoir\'{e}ing Benchmark \\ from Synthesis to Restoration}

\titlerunning{Fabric Image Demoir\'{e}ing Benchmark}

\author{Pengchao Wei \and
Xiaojie Guo\textsuperscript{\Letter}}

\authorrunning{P.~Wei and X.~Guo}

\institute{
Tianjin University, Tianjin, China\\
\email{\{pcwei.ai,xj.max.guo\}@gmail.com} \\
Project page: \url{https://weipengchao.top/PRISM-page}
}

\maketitle

\begingroup
\renewcommand{\thefootnote}{}
\footnotetext{\Letter~indicates corresponding author.}
\endgroup

\begin{abstract}
Fabric moir\'{e} is a sampling-induced aliasing artifact caused by the interaction between fine textile patterns and camera sensor grids, producing structured interference that severely degrades image quality. Unlike screen-induced moir\'{e}, which stems from strictly periodic display lattices, fabric moir\'{e} is intrinsically more challenging due to the broadband and semi-periodic nature of textile weaves. The heavy spectral overlap between intrinsic texture and aliasing components renders fabric demoir\'{e}ing substantially more ill-posed. Consequently, existing models trained on screen moir\'{e} datasets generalize poorly to these complex textile patterns.
Despite its practical importance, fabric image demoir\'{e}ing remains underexplored and lacks standardized benchmarks. We present the first comprehensive benchmark for fabric image demoir\'{e}ing. To address the difficulty of acquiring pixel-aligned real-world pairs, we develop a physically motivated synthesis framework and construct a large-scale dataset comprising 16,050 paired multi-resolution fabric images with controllable aliasing severity. Furthermore, we customize a baseline model, which establishes promising performance on the proposed benchmark dataset with strong generalization ability. Our benchmark provides a standardized platform for advancing research in fabric image demoir\'{e}ing. 
\end{abstract}

\section{Introduction}
\label{sec:intro}

When a discrete imaging array samples high-frequency scene content below the Nyquist–Shannon limit \cite{shannon2006communication,nyquist1928certain}, spectral folding occurs, causing high-frequency components to alias into the baseband. The resulting interference manifests as structured patterns, such as periodic stripes, ripples, and chromatic distortions, commonly referred to as moir\'{e} artifacts. Formally, moir\'{e} arises from spectral overlap between intrinsic scene frequencies and sampling-induced replicas. When this overlap intersects with genuine image content, signal and artifact become spectrally entangled, rendering inverse recovery fundamentally ill-posed.

Fabrics are among the most challenging real-world sources of moir\'{e}. Textile weaves contain dense, anisotropic, and semi-periodic micro-patterns (\eg, fine stripes, grids, herringbone, and tight weaves), whose spectra are broadband and often approach the sensor’s Nyquist boundary. Unlike screen-induced moir\'{e}, which originates from strictly periodic display lattices with discrete spectral peaks, fabric textures have distributed frequency support. Consequently, aliasing overlaps heavily with intrinsic texture, producing strong spectral entanglement and substantially increasing restoration ambiguity. Additionally, fabrics are non-rigid surfaces that exhibit wrinkles, deformations, and pose variations, further complicating demoir\'{e}ing.

\begin{figure}[t]
  \centering
  \includegraphics[width=\linewidth]{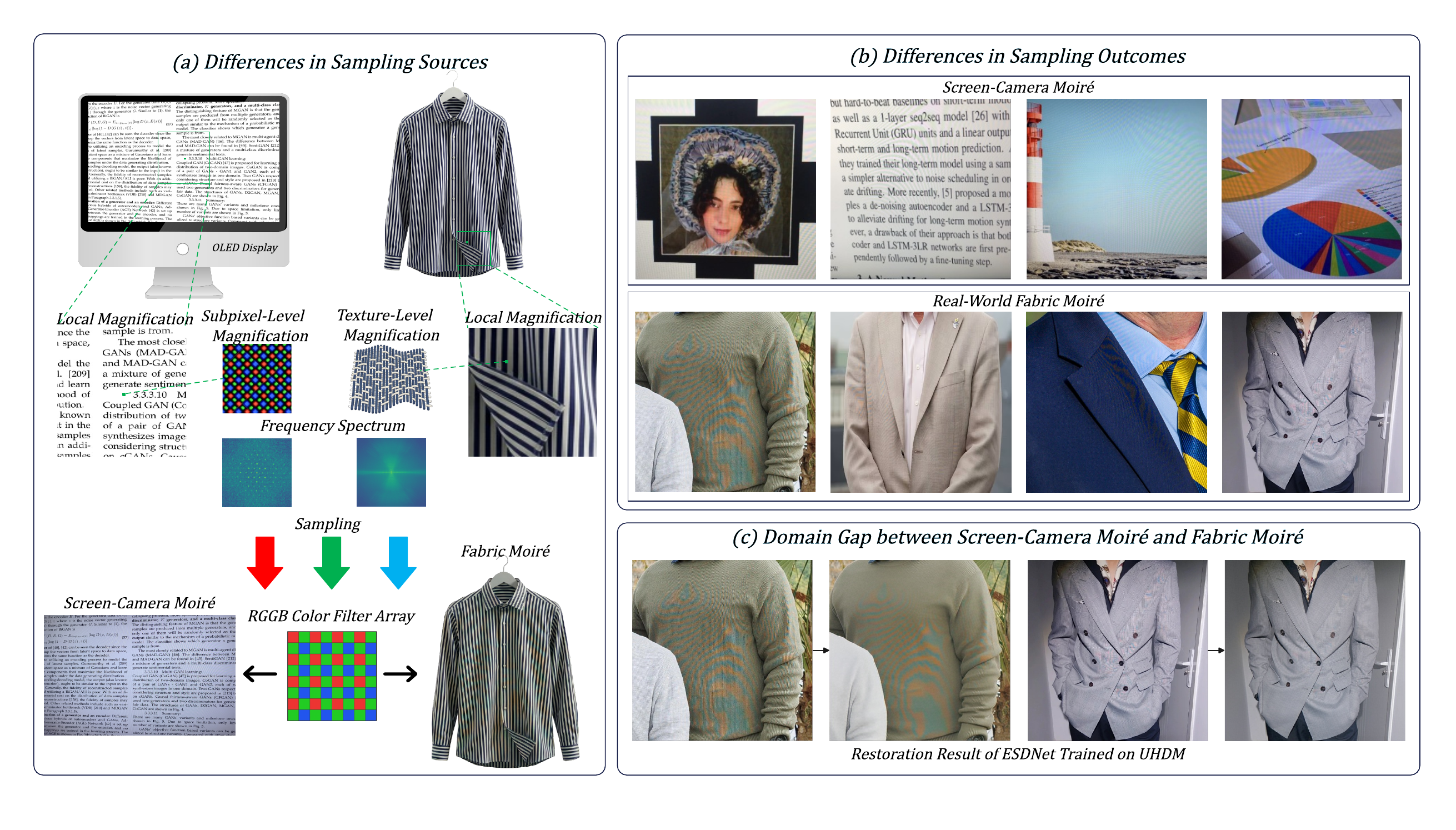}
  \caption{\textbf{Cause-to-appearance domain gap between screen moir\'{e} and fabric moir\'{e}.} (a) Sampling sources: screen moir\'{e} originates from rigid planar displays with regular pixel/subpixel lattices, while fabric moir\'{e} arises on non-rigid, deformable garments with anisotropic weave microstructures, undergoing folds, stretching, and pose variations. (b) Sampling outcomes: screen moir\'{e} is typically global and content-agnostic, whereas fabric moir\'{e} is localized to high-frequency textile regions and strongly coupled with texture and deformation. (c) Domain gap: these differences hinder transferring screen-trained models, leading to moir\'{e} residual or over-smoothing.}
  \label{fig:contrast}
  \vspace{-10pt}
\end{figure}

Recent learning-based demoir\'{e}ing methods have made significant progress in screen-camera scenarios, owing to paired datasets and architectures designed for periodic display patterns. Early works leveraged multi-scale decompositions to address the cross-frequency characteristics of moir\'{e} \cite{sun2018moire,cheng2019multi}, while subsequent methods incorporated structural priors and frequency-domain modeling, such as feature fusion for moir\'{e} attributes \cite{he2019mop}, wavelet-domain processing \cite{liu2020wavelet}, and multi-band modeling \cite{zheng2020image}. While in high-resolution settings, designs focusing on efficiency and scale robustness have also emerged \cite{he2020fhde2net,yu2022towards,xiao2024p,lee2025moir}. However, these methods assume spectral separability between content and artifact, which fails in fabrics. Directly transferring screen-oriented models often results in over-smoothing, texture removal, or incomplete artifact suppression (see \cref{fig:contrast}, which shows examples of removing fabric moiré using UHDM-trained ESDNet\cite{yu2022towards}), demonstrating a fundamental domain gap. 
Moreover, supervised demoir\'{e}ing is often sensitive to pixel-level alignment. Misalignment between moir\'{e} input images and their corresponding clean targets may cause unstable optimization and significantly degrade restoration quality. Even in the ``screen is flat'' ideal condition, constructing pixel-aligned data requires complex registration engineering. For example, the TIP2018 dataset designs high-contrast black borders and corner structures around the reference images and performs registration using homography~\cite{sun2018moire}; the UHDM 4K dataset employs RANSAC~\cite{fischler1981random} to estimate homography and manual selection to remove severely misaligned samples~\cite{yu2022towards}. In addition, the FHDMi dataset acknowledges 5--10 pixel shifts caused by nonlinear camera distortions and adopts a learning-based tolerance strategy by employing the Contextual Bilateral (CoBi) loss \cite{zhang2019zoom} to match features within a local neighborhood~\cite{he2020fhde2net}. More recent work further demonstrates that additional feature-matching-based alignment (such as SIFT~\cite{lowe2004distinctive}) on UHDM considerably improves training stability and restoration quality, highlighting the decisive role of alignment in demoir\'{e}ing tasks~\cite{xiao2024p}. While for fabric scenes, this alignment issue is further magnified, because non-rigid deformations and subtle pose changes make global alignment infeasible, thus preventing practical large-scale collection. 

\begin{figure}[t]
  \centering
  \includegraphics[width=\linewidth]{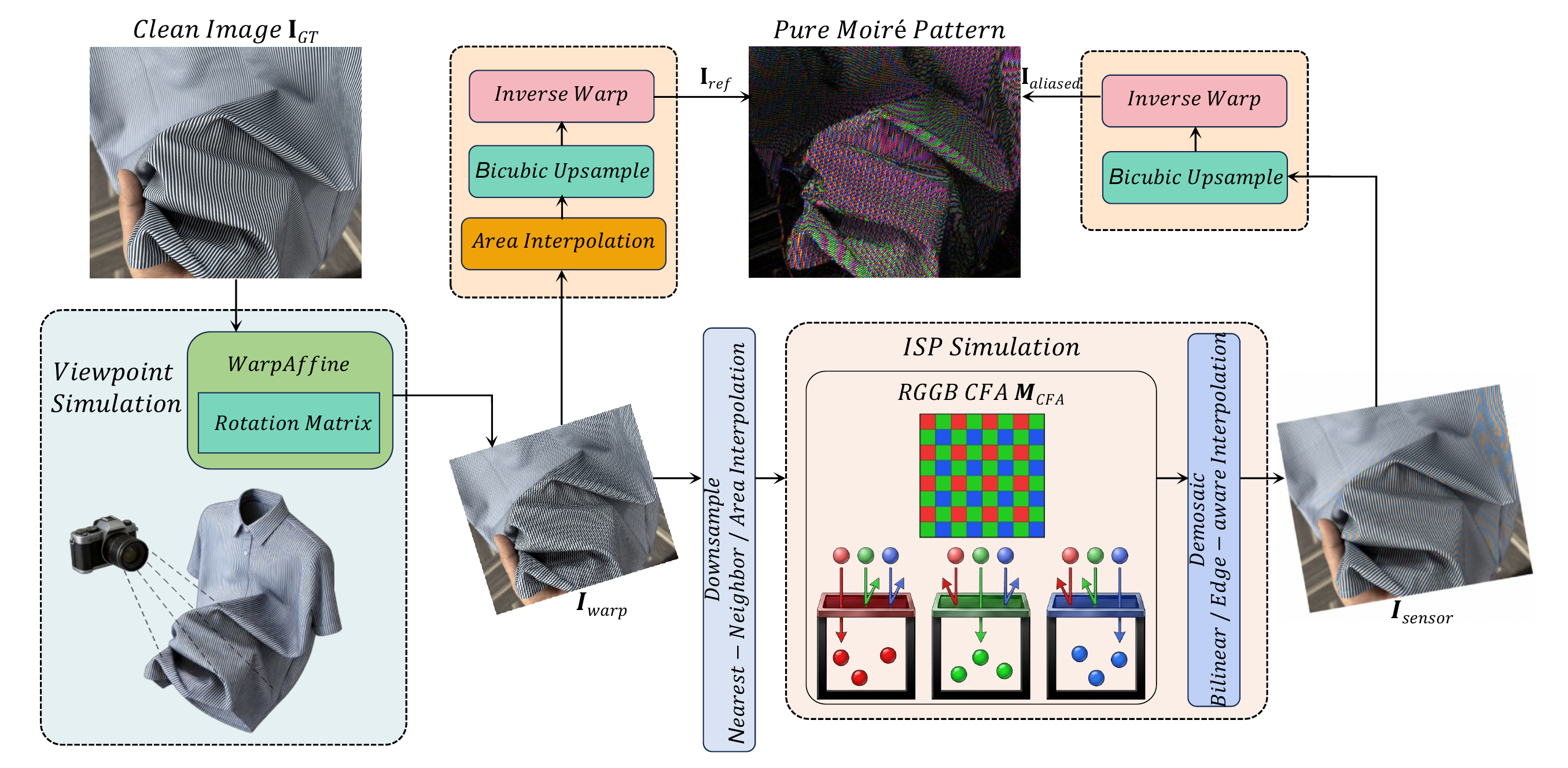}
  \caption{Our synthesis pipeline}
   \vspace{-14pt}
  \label{fig:syn_pipeline}
\end{figure}

To overcome the above challenge, we introduce PRISM (Physics-based Residual Injection for Synthetic Moir\'{e}), a synthesis framework that models the imaging chain and generates paired fabric moir\'{e} with precise pixel-level alignment. 
PRISM extracts aliasing residuals via a round-trip construction and injects them into the original pixel grid, preserving genuine textile textures while simulating realistic aliasing. Using a one-to-many strategy, we generate a large-scale multi-resolution dataset of 16,050 fabric images covering diverse levels of aliasing. Additionally, a small real-world unpaired dataset is also collected to validate zero-shot generalization.
Besides, building on the PRISM benchmark dataset, we further customize a fabric demoir\'{e} network, namely FaDeNet, to faithfully restore fabrics. FaDeNet explicitly decomposes the input into low-frequency base and high-frequency detail components. The low-frequency branch uses a multi-scale U-shaped trunk\cite{ronneberger2015u} to predict residual corrections and a spatial confidence mask, and applies corrections only in the required regions using a mask-gated update rule. In parallel, a lightweight detail branch performs magnitude-bounded refinement on the high-frequency component under the same mask to suppress moir\'{e} interference while avoiding excessive modification of textile textures. Furthermore, we propose Spectral-Anisotropic Gated Blocks (SAGB) to capture anisotropic stripe structures in the spatial domain and narrow-band periodic cues in the frequency domain, enabling adaptive correction strength in moir\'{e}-dominant regions while suppressing unnecessary modifications in clean textures. 

Our major contributions are summarized as follows:
\begin{itemize}
\item \textbf{First dedicated fabric demoir\'{e} benchmark.}
We systematically investigate fabric image demoir\'{e}ing in real-world photography and introduce PRISM, a physics-based residual injection synthesis framework that generates realistic fabric moir\'{e} while guaranteeing perfect pixel-level alignment and preserving intrinsic textile textures. Based on PRISM, we construct a large-scale, multi-resolution paired dataset and validate its realism and transferability through zero-shot evaluation on real unpaired fabric images. 

\item \textbf{Benchmarking existing methods and a fabric-tailored model.}
We benchmark representative screen-camera demoir\'{e} architectures on PRISM and further introduce FaDeNet, a conservative restoration network with base/detail decomposition, mask-gated low-frequency correction, magnitude-bounded detail refinement, and spectral-anisotropic gated blocks. FaDeNet achieves state-of-the-art performance on PRISM across PSNR, SSIM, and LPIPS, with low computational overhead, establishing competitive and reproducible baselines for future research.
\end{itemize}

\section{Related Work}
\subsection{Screen-Camera Image Demoir\'{e}ing} 
Moir\'{e} artifacts in screen-camera imaging arise from frequency aliasing between the display subpixel lattice and the camera sampling process. Before end-to-end learning became dominant, demoir\'{e}ing was mostly studied through hand-crafted priors. Sidorov and Kokaram~\cite{sidorov2002suppression} proposed suppressing telecine-related moir\'{e} via spectral analysis. Sur and Gr\'{e}diac~\cite{sur2015automated} addressed periodic/quasi-periodic noise using frequency-domain statistics and automated notch-filter design. Liu et al.~\cite{liu2015moire} formulated moir\'{e} removal as a patch-wise layer decomposition, regularizing textures with low-rank priors. Yang et al.~\cite{yang2017textured} further studied textured image demoir\'{e}ing via signal decomposition and guided filtering. Meanwhile, descreening methods for scanned halftone prints (\eg, training-based descreening~\cite{siddiqui2007training} and wavelet-based filtering~\cite{luo1998robust}) also constitute an important line of early exploration. Subsequent learning-based methods typically adopt supervised training on paired moir\'{e}/clean images. Sun et al.~\cite{sun2018moire} introduced DMCNN, explicitly leveraging multi-resolution feature maps to remove moir\'{e} artifacts across different frequency bands. Later works further improved restoration quality through dynamic multi-scale feature encoding \cite{cheng2019multi}, property-oriented demoir\'{e}ing with moir\'{e}-type cues~\cite{he2019mop}, and frequency-aware modeling. In particular, Zheng et al.~\cite{zheng2020image} formulated demoir\'{e}ing as the combination of moir\'{e} texture removal and color/tone correction, and introduced learnable bandpass filters to better capture moir\'{e}-related frequency priors. Liu et al.~\cite{liu2020wavelet} further exploited wavelet decomposition for targeted artifact suppression. To handle 4K resolutions, He et al.~\cite{he2020fhde2net} proposed FHDe$^2$Net, a cascaded global-to-local pipeline. Yu et al.~\cite{yu2022towards} introduced the UHDM dataset and the lightweight ESDNet with scale-aware fusion. For UHD deployment, Xiao et al.~\cite{xiao2024p} developed P-BiC for patch-level restoration, while Lee et al.~\cite{lee2025moir} explored large-kernel CNNs (MZNet) to capture long-range structures. Recent works also explore RAW-domain demoir\'{e}ing for screen recapture, showing that RAW observations can provide useful information before ISP-induced color and frequency mixing~\cite{yue2022recaptured,xu2024image}. In contrast to these screen-camera studies, we investigate fabric demoir\'{e}ing and introduce PRISM and FaDeNet for texture-faithful restoration.

\subsection{Fabric Demoir\'{e}ing}
While various moir\'{e} synthesis schemes have been proposed~\cite{yang2025unidemoire, zhong2024learning, park2022unpaired, yuan2019aim, yang2023doing}, they are mainly designed for screen-camera degradations, where moir\'{e} arises from the aliasing between display sub-pixel grids and camera sampling processes. These methods typically rely on the implicit assumption that moir\'{e} can be modeled as a content-independent foreground layer. However, this assumption is physically implausible for textiles, as fabric moir\'{e} is coupled with the intrinsic textile micro-structures. Consequently, demoir\'{e}ing models optimized for screen-camera scenarios tend to over-smooth authentic textures or fail to suppress interference that is tightly entangled with the textile geometry. To our knowledge, research specifically addressing fabric moir\'{e} remains sparse. The most relevant work is by Liu et al.~\cite{liu2020self}, which leverages focused/defocused observations for learning and highlights that screen-oriented data pipelines are ill-suited for high-texture demoir\'{e}ing. Nevertheless, their approach necessitates a specialized hardware setup and a rigid acquisition pipeline to obtain paired inputs, posing significant challenges for practical data collection at scale. Compared with the above methods, we aim to enable a stable and scalable paired supervision for fabric demoir\'{e}ing.

\section{PRISM: Benchmark Dataset Construction}
\label{sec:dataset}

\begin{figure}[t]
  \centering
  \includegraphics[width=\linewidth]{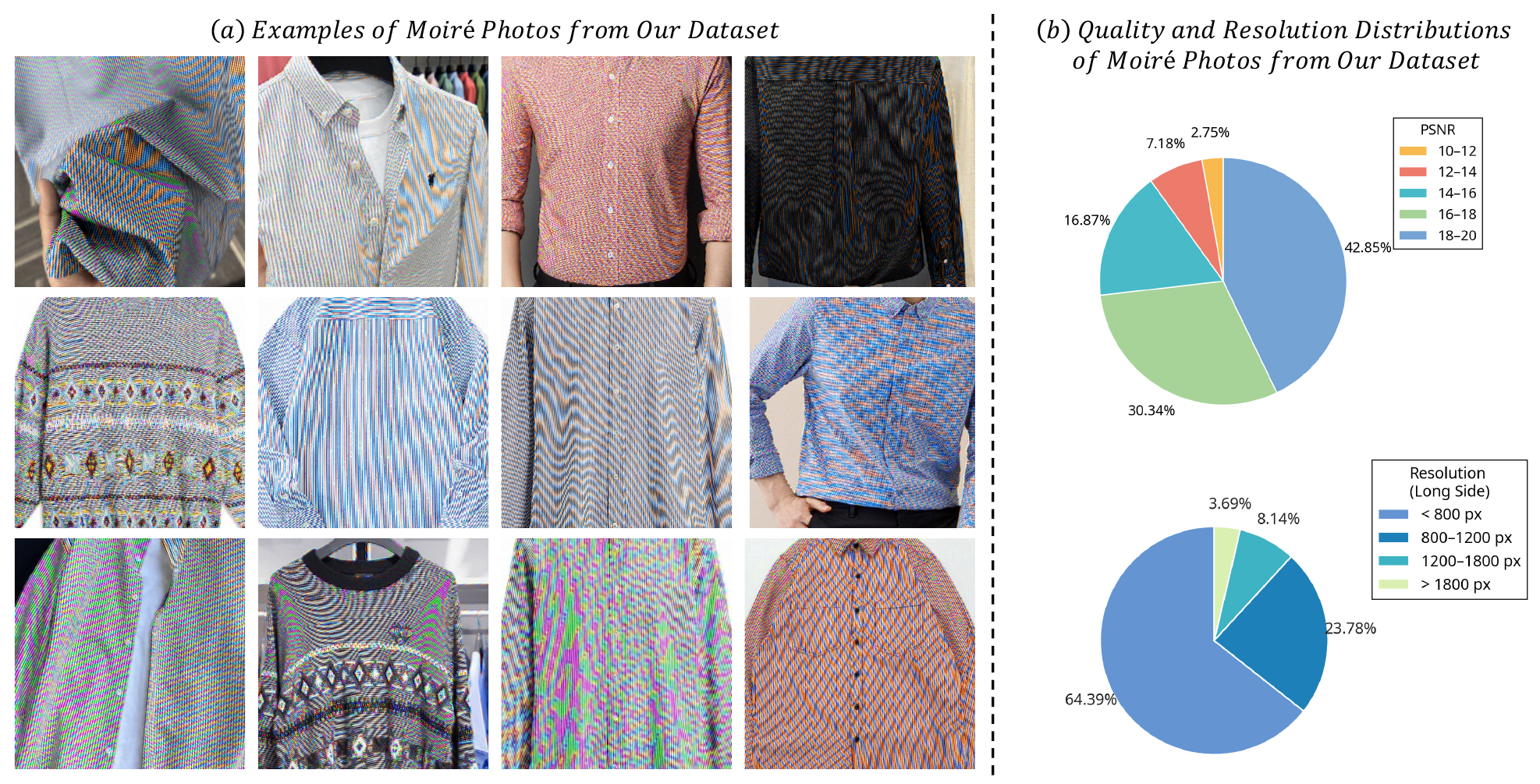}
  \caption{Overview of the PRISM dataset. (a) Representative synthesized moir\'{e} images in PRISM. (b) Distributions of image quality and resolution.}
  \label{fig:prism_dataset}
  \vspace{-12pt}
\end{figure}

We investigate fabric image demoiréing under paired supervision. As illustrated in \cref{fig:contrast}, due to significant differences in the physical microstructures of the sampling sources, the data distribution of screen-camera moiré diverges from that of fabric moiré. Specifically, fabric moiré patterns are more localized within the highly textured regions of fabric and exhibit greater pattern complexity. They tend to distort, degrade, or become heavily coupled with the intrinsic textures of the fabric, making them more difficult to remove. 
Taking ESDNet~\cite{yu2022towards} as an example, when the model is trained on the screen-camera dataset UHDM \cite{yu2022towards}, it fails to effectively remove fabric moiré patterns and often suffers from unintended eradication of fabric textures. Therefore, a dedicated fabric moiré dataset is necessary. However, acquiring large-scale, pixel-aligned real-world fabric moiré datasets is costly and technically challenging. 

From an alignment perspective, factors such as non-linear lens distortion, capture perturbations, and the interference of moiré patterns themselves introduce persistent alignment errors. In datasets like TIP18~\cite{sun2018moire}, FHDMi~\cite{he2020fhde2net}, and UHDM, even with sophisticated registration engineering as we discussed in \cref{sec:intro}, noticeable alignment issues still exist in the data. 
Moreover, from a data acquisition perspective, screen-camera moir\'e pairs can be collected by displaying Internet images on a monitor and re-photographing the screen, where the original images serve as GT after standard processing. In contrast, fabric moiré presents a unique challenge as it must be captured from physical garments with high-frequency textures. The primary bottleneck lies in the inability to obtain perfectly aligned, moiré-free ground truth (GT) images from the identical camera perspective. This lack of pixel-perfect pairs makes the acquisition of high-quality training data a difficult task, significantly hindering the application of standard supervised learning methods.
Motivated by these constraints, we propose PRISM (Physics-based Residual Injection for Synthetic Moiré) and construct a large-scale paired synthetic dataset for stable supervised learning.

\subsection{Data Collection and Pair Synthesis}
To better match real demoir\'{e}ing scenarios, we restrict fabric materials to high-frequency textures prone to moir\'{e} and manually remove samples with pre-existing artifacts. Under these constraints, we collect $\sim$2,000 fabric images from the Internet as GT. Based on the synthesis pipeline depicted in \cref{fig:syn_pipeline}, we adopt a one-to-many strategy to better simulate moiré variations under diverse real-world capture conditions. In our release, PRISM contains 16,050 synthesized multi-resolution fabric moir\'{e} pairs.

\subsection{PRISM Synthesis Pipeline}
\label{sec:syn}

\subsubsection{Problem Formulation}
The general synthesis workflow is illustrated in \cref{fig:syn_pipeline}. Given a clean fabric image $\mathbf{I}_\text{GT}\in\mathbb{R}^{H\times W\times 3}$, PRISM aims to synthesize a moir\'{e}-corrupted observation while maintaining pixel-wise alignment with $\mathbf{I}_\text{GT}$.

We derive PRISM from a physics-based digital imaging chain. In real captures, moir\'{e} artifacts are shaped by multiple factors: (i) capture geometry (viewpoint, in-plane rotation, shooting distance) that changes the relative scale and orientation between the CFA sampling lattice and the sampled objects; (ii) discrete sensor sampling that can undersample high-frequency object content beyond the sensor’s Nyquist limit\cite{nyquist1928certain}, causing spectral folding; and (iii) sensor photoelectric effects and ISP (CFA mosaicking, sensor noise, demosaicking\cite{malvar2004high}, and color rendering) that modulate the visual appearance of moir\'{e}, especially chromatic fringes. A simplified imaging chain can be written as
\begin{equation}
    \mathbf{I}[\mathbf{n}] = \mathcal{R}_{\text{ISP}}\left(\mathcal{S}_{\Delta}\left(\mathcal{T}_{\theta,s}\big(L(\mathbf{x})\big)\right)\right)+\eta,
    \label{eq:image_chain}
\end{equation}
where $\mathbf{I}[\mathbf{n}]$ is the discrete RGB image sampled on the pixel lattice indexed by $\mathbf{n}\in\mathbb{R}^2$, $L(\mathbf{x})$ denotes the continuous light field defined over spatial coordinate $\mathbf{x}\in\mathbb{R}^2$, $\mathcal{T}_{\theta,s}$ is a geometric projection parameterized by observation angle $\theta$ and distance $s$, $\mathcal{S}_{\Delta}$ is a discretization operator that samples and integrates the projected signal onto a sensor grid with pitch $\Delta$, $\mathcal{R}_{\text{ISP}}$ denotes an ISP-like rendering pipeline, $\eta$ models photoelectronic noise.

Although \cref{eq:image_chain} characterizes the physical formation of moir\'{e} patterns, directly employing this process for image synthesis leads to an irreversible loss of high-frequency details due to the low-pass filtering nature of the discretization operator $\mathcal{S}_{\Delta}$. Such degradation hampers the model to learn the restoration of fine-grained textures. To provide the high-fidelity texture references, PRISM avoids reconstructing the entire degraded signal field. Instead, it adopts the residual injection strategy as:
\begin{equation}
\mathbf{I}_{\text{Moire}}=\mathbf{I}_\text{GT}+\gamma\cdot \Delta\boldsymbol{\Phi}(\mathbf{I}_\text{GT};\theta,s),
\label{eq:prism}
\end{equation}
where $\gamma$ controls the injection strength. 
By superimposing the physical moiré residual field $\Delta\boldsymbol{\Phi}$ onto $\mathbf{I}_\text{GT}$, we ensure that the underlying texture integrity is perfectly preserved. 
This strategy constructs an ideal training pair of clean texture, forcing the network to precisely decouple artifacts from complex backgrounds.

\vspace{-8pt}
\subsubsection{Imaging-Chain Simulation}
We generate moir\'{e} patterns by simulating the camera capture process, which includes the viewpoint relationship relative to the object and the subsequent ISP pipeline.

\paragraph{Viewpoint Simulation.} We simulate the viewpoint relationship between the fabric and the sensor through a geometric manifold projection $\mathcal{T}_{\theta, s}$.
\begin{equation}
    \mathbf{I}_{\text{warp}} = \mathcal{T}_{\theta, s}(\mathbf{I}_\text{GT}).
\end{equation}
where the observation angle $\theta$ is sampled uniformly from $[-25^\circ, 25^\circ]$ and the distance factor $s$ follows a Beta distribution $\mathcal{B}(1.2, 1.2)$ rescaled to $[0.25, 0.85]$.

Following the geometric warping, we apply a discrete sampling operator $\mathcal{S}_{\downarrow}$ to induce spectral folding:
\begin{equation}
\mathbf{I}_{\text{tiny}} = \mathcal{S}_{\downarrow}(\mathbf{I}_{\text{warp}}).
\label{eq:sampling}
\end{equation}

To further approximate the real ISP, $\mathbf{I}_{\text{tiny}}$ is subsampled according to an RGGB Bayer mosaic $\mathbf{M}_{\text{CFA}}$ with additive photon shot noise $\mathcal{N}(0,\sigma^2)$:

\begin{equation}
\mathbf{R}_{\text{raw}}=(\mathbf{I}_{\text{tiny}}\odot \mathbf{M}_{\text{CFA}})+\mathcal{N}(0,\sigma^2).
\label{eq:cfa_noise}
\end{equation}

\paragraph{ISP Simulation.}
Finally, a demosaic operator $\mathcal{R}_{\text{dm}}(\cdot)$ is employed. As implemented in our PRISM framework, we randomly select between standard bilinear and edge-aware interpolation to yield the RGB sensor observation $\mathbf{I}_{\text{sensor}}$:
\begin{equation}
\mathbf{I}_{\text{sensor}} = \mathcal{R}_{\text{dm}}(\mathbf{R}_{\text{raw}}).
\label{eq:isp}
\end{equation}
This physics-based pipeline ensures that the synthesized moir\'{e} patterns exhibit realistic chromatic fringes and structural distortions.

\subsubsection{Round-trip Alignment}
The conventional forward imaging chain (Eq.~\ref{eq:image_chain}) is inherently destructive, as the discrete sampling operator $\mathcal{S}_{\Delta}$ irreversibly filters out textile details. To circumvent this, PRISM isolates pure moir\'{e} artifacts by contrasting two geometry-matched branches.

\paragraph{Aliased path.}
We upsample the rendered sensor output and map it back to the canonical high-resolution coordinate system:
\begin{equation}
    \mathbf{I}_{\text{aliased}} = \mathcal{T}^{-1}_{\theta,s} \left( \text{interp}_{\uparrow}(\mathbf{I}_{\text{sensor}}) \right),
    \label{eq:aliased}
\end{equation}
where $\text{interp}_{\uparrow}$ is implemented via bicubic upsampling.
\paragraph{Reference Path.}
As an anti-aliased reference, we apply Nyquist-compliant downsampling $\mathcal{S}_{\downarrow}^{\text{LPF}}$ via area interpolation to $\mathbf{I}_{\text{warp}}$:
\begin{equation}
    \mathbf{I}_{\text{ref}} = \mathcal{T}^{-1}_{\theta,s} \left( \text{interp}_{\uparrow} \big( \mathcal{S}_{\downarrow}^{\text{LPF}}(\mathbf{I}_{\text{warp}}) \big) \right).
    \label{eq:ref}
\end{equation}
By keeping the same operators as \cref{eq:aliased}, $\mathbf{I}_{\text{ref}}$ effectively isolates the bias introduced by pure geometric resampling without introducing aliasing components.

\paragraph{Moir\'{e} extraction.} 
The pure moir\'{e} residual field $\Delta\boldsymbol{\Phi}$ is extracted by:
\begin{equation}
    \Delta\boldsymbol{\Phi}(\mathbf{I}_\text{GT}; \theta, s) = \mathbf{I}_{\text{aliased}} - \mathbf{I}_{\text{ref}}.
    \label{eq:residual}
\end{equation}

\section{FaDeNet: Proposed Fabric Demoir\'{e}ing Network}
\label{sec:method}

\begin{figure}[t]
  \centering
  \includegraphics[width=\linewidth]{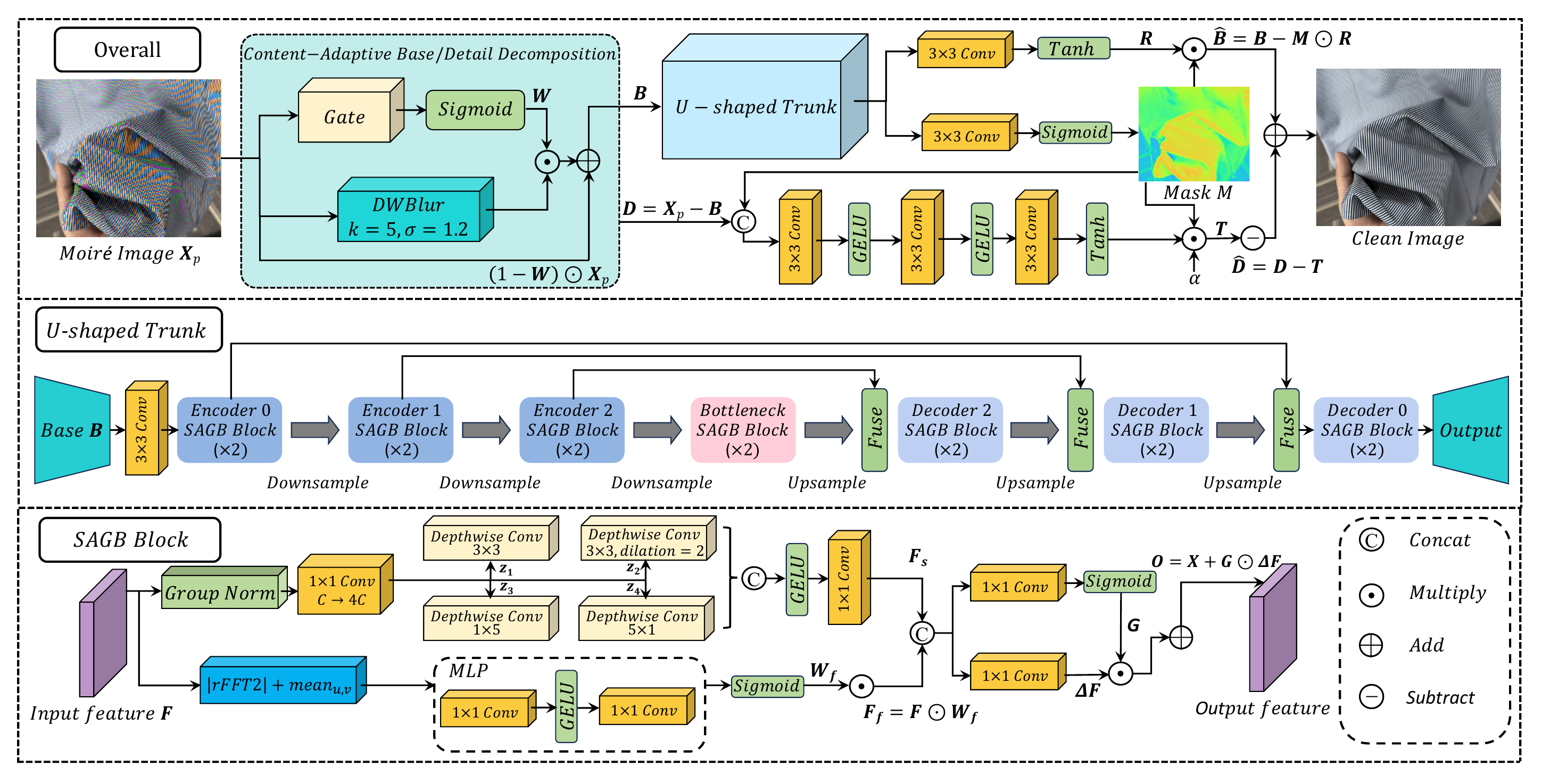}
  \caption{Pipeline of our method.}
  \label{fig:pipeline}
\end{figure}

\vspace{-10pt}

We propose FaDeNet for fabric demoir\'{e}ing. As illustrated in \cref{fig:pipeline}, the overall architecture starts with a content-adaptive base/detail decomposition, followed by two correction branches with a shared spatial confidence mask.

Given an input moir\'{e} image $\mathbf{X}\in\mathbb{R}^{H\times W\times 3}$, we first pad it to a multiple of $8$ for pixel-rearrangement down/up-sampling:
$\mathbf{X}_p=\mathrm{Pad}_8(\mathbf{X})$.
We then compute a content-adaptive base/detail split $(\mathbf{B},\mathbf{D})$, where the base captures low-frequency variations (\eg, color/illumination shifts) and the detail keeps high-frequency structures (\eg, edges/textures).

Our 4-scale U-shaped trunk operates only on the base component and predicts (i) a base residual $\mathbf{R}$ and (ii) a spatial confidence mask $\mathbf{M}\in(0,1)^{H\times W\times 1}$. Scale changes are implemented by PixelUnshuffle/PixelShuffle ($\times2$)\cite{shi2016real} to reduce aliasing. The base is updated by a masked conservative rule:
\begin{align}
\hat{\mathbf{B}} = \mathbf{B} - \mathbf{M}\odot \mathbf{R}.
\label{eq:base_update}
\end{align}
This enforces corrections only where the mask indicates moir\'{e}-dominant regions, improving stability and avoiding over-modification in clean areas.

To mitigate over-smoothing, we further apply a lightweight, mask-gated and magnitude-controlled correction on the high-frequency detail:
\begin{align}
\hat{\mathbf{D}} = \mathbf{D} - \alpha\,\mathbf{M}\odot \tanh(\mathcal{R}([\mathbf{D},\mathbf{M}])),
\label{eq:detail_update}
\end{align}
where $\mathcal{R}(\cdot)$ is a small residual predictor, and $\alpha$ is an empirically selected magnitude bound for controlling the detail correction strength. We set $\alpha{=}0.35$ by default to balance moir\'{e} suppression and texture preservation.

Finally, the restored output is recomposed and cropped back:
\begin{align}
\hat{\mathbf{Y}}=\mathrm{Crop}\big(\hat{\mathbf{B}}+\hat{\mathbf{D}}\big).
\label{eq:recompose}
\end{align}
%
\subsection{Content-Adaptive Base/Detail Decomposition}
\label{sec:decomp}
Moir\'{e} artifacts typically manifest as low-frequency color shifts coupled with high-frequency periodic interference. Instead of a fixed blur, we employ a content-adaptive depthwise Gaussian smoothing to split frequency bands:
\begin{align}
\mathbf{W} = \sigma(\mathrm{Gate}(\mathbf{X}_p)), \  
\mathbf{B} = \mathbf{W}\odot \mathrm{Blur}(\mathbf{X}_p) + (1-\mathbf{W})\odot \mathbf{X}_p, \ 
\mathbf{D} = \mathbf{X}_p - \mathbf{B}, \label{eq:detail_def}
\end{align}
where $\mathbf{W}$ is a single-channel gating map; $\mathrm{Blur}(\cdot)$ is depthwise Gaussian conv~\cite{howard2017mobilenets} with $(k{=}5,\sigma{=}1.2)$ and $\mathrm{Gate}(\cdot)$ is a lightweight convolution stack.

\subsection{Spectral-Anisotropic Gated Block (SAGB)}
\label{sec:sagb}

Moir\'{e} patterns typically manifest as anisotropic spatial stripes and narrow-band frequency peaks. SAGB is designed to capture both with low overhead.

\subsubsection{Spatial anisotropic branch.}
Let $\mathbf{F}$ denote the input feature map of SAGB. Given the normalized features $\tilde{\mathbf{F}} = \mathrm{GN}(\mathbf{F})$, we apply a $4C$ expansion followed by multi-branch anisotropic depthwise filtering, and finally fuse the features by:

\begin{align}
\mathbf{F}_s
&= \mathcal{P}\,\!\Big(
\phi\big([\mathrm{DW}_{3\times3}(\mathbf{Z}_1),\,
\mathrm{DW}^{d=2}_{3\times3}(\mathbf{Z}_2),\,
\mathrm{DW}_{1\times5}(\mathbf{Z}_3),\,
\mathrm{DW}_{5\times1}(\mathbf{Z}_4)]\big)\Big),
\label{eq:sagb_spatial}
\end{align}
where $[\mathbf{Z}_1,\mathbf{Z}_2,\mathbf{Z}_3,\mathbf{Z}_4]$ is a channel-wise split of the expanded feature $\mathrm{Conv}_{1\times1}(\tilde{\mathbf{F}})$ (from $C$ to $4C$), $\mathcal{P}$ denotes point-wise ($1\times1$) conv and $\phi$ denotes GELU\cite{hendrycks2016gaussian}.

\subsubsection{Spectral gate}
Prior work has shown that leveraging frequency-domain representations can benefit image restoration\cite{chi2020fast, jiang2024fast}. To sense periodic interference, we compute a channel-wise gate from the 2D real FFT magnitude~\cite{bracewell1986fourier}:

\begin{align}
\mathbf{W}_f = \sigma\!\Big(\mathrm{MLP}\big(\mathrm{mean}_{u,v}(|\mathrm{rFFT2}(\mathbf{F})|)\big)\Big), \quad
\mathbf{F}_f = \mathbf{F}\odot \mathbf{W}_f .
\label{eq:sagb_fft}
\end{align}
For efficiency, the spectral gate is applied only at coarse scales (1/4, 1/8).

\subsubsection{Gated residual update.}
We concatenate $\mathbf{C}=[\mathbf{F}_s,\mathbf{F}_f]$ and predict an update:
\begin{align}
\mathbf{G}=\sigma(\mathrm{Conv}_{1\times1}(\mathbf{C})),\quad
\Delta\mathbf{F}=\mathrm{Conv}_{1\times1}(\mathbf{C}),\quad
\mathbf{O}=\mathbf{F}+\mathbf{G}\odot \Delta\mathbf{F}.
\end{align}

This formulation allows stronger corrections in moir\'{e}-dominant regions while suppressing unnecessary updates in clean, texture-rich areas.

\subsection{Loss Function}
\label{sec:loss}
We minimize an $\ell_1$ loss with a Laplacian high-pass constraint~\cite{marr1980theory,sobel19683x3}:
\begin{align}
\mathcal{L}=\|\hat{\mathbf{Y}}-\mathbf{Y}\|_1
+\lambda_{\mathrm{hp}}\cdot\|\mathrm{Lap}(\hat{\mathbf{Y}})-\mathrm{Lap}(\mathbf{Y})\|_1,
\label{eq:loss}
\end{align}
where $\lambda_{\mathrm{hp}}=0.3$ is set empirically, and $\mathrm{Lap}(\cdot)$ denotes a Laplacian operator.

\section{Experiments}
\label{sec:exp}

\subsection{Experimental Setting}

\subsubsection{Datasets and Metrics.}
We perform benchmark evaluations on the proposed PRISM dataset, and we report PSNR, SSIM~\cite{wang2004image} and LPIPS~\cite{zhang2018unreasonable} for quantitative comparison. 
In addition to restoration quality, we further report efficiency metrics, including computational cost (GFLOPs), model size (Params), and inference throughput (FPS) on PRISM. 
To assess real-world robustness, we further evaluate zero-shot transfer on unpaired fabric captures and report no-reference perceptual metrics on the real-world fabric moir\'{e} set, including ARNIQA~\cite{agnolucci2024arniqa} and CLIP-IQA+ (ViT-L/14, 512)~\cite{wang2023exploring,radford2021learning}, complemented by human mean opinion score (MOS) ratings. 

\subsubsection{Implementation Details.}
All experiments were conducted on a single NVIDIA RTX 4090 GPU. During training, we randomly cropped $384\times384$ patches with a batch size of 4 and optimized the model for 200 epochs using Adam~\cite{kingma2014adam} ($\beta_1{=}0.9$, $\beta_2{=}0.999$). We set the initial learning rate to $2\times10^{-4}$ and applied a cyclic cosine annealing schedule~\cite{loshchilov2016sgdr}. For efficiency comparisons, we benchmarked all methods using a fixed-resolution input of $384\times384$. All compared methods were retrained from scratch on the PRISM dataset to ensure their optimal performance.

\begin{table}[t]
  \centering
  \caption{Performance and efficiency comparison on PRISM. Bold and underlined values indicate the best and second-best results for PSNR/SSIM/LPIPS, respectively.}
  \label{tab:prism_quant}
  \setlength{\tabcolsep}{4.5pt}
  \renewcommand{\arraystretch}{1.08}
  \resizebox{\linewidth}{!}{
  \begin{tabular}{llccc ccc}
    \toprule
    \multirow{2}{*}{Method} & \multirow{2}{*}{Venue} &
    \multicolumn{3}{c}{Quality} &
    \multicolumn{3}{c}{Efficiency} \\
    \cmidrule(lr){3-5}\cmidrule(lr){6-8}
    & & PSNR $\uparrow$ & SSIM $\uparrow$ & LPIPS $\downarrow$ & GFLOPs  & Params (M) & FPS \\
    \midrule
    Input                            & --         & 16.998 & 0.7268 & 0.3495 & --      & --     & --     \\
    DMCNN\cite{sun2018moire}         & TIP 2018   & 24.659 & 0.9307 & 0.0772 & 17.648  & 1.426  & 164.85 \\
    MDDM\cite{cheng2019multi}        & ICCVW 2019 & 24.916 & 0.9260 & 0.0755 & 57.489  & 8.009  & 17.74  \\
    WDNet\cite{liu2020wavelet}       & ECCV 2020  & 26.141 & 0.9444 & 0.0524 & 34.195  & 3.919  & 65.00  \\
    MBCNN\cite{zheng2020image}       & CVPR 2020  & 27.911 & 0.9617 & 0.0405 & 133.754 & 14.192 & 68.41  \\
    MopNet\cite{he2019mop}           & ICCV 2019  & 28.190 & 0.9648 & 0.0362 & 408.037 & 60.228 & 10.05  \\
    FHDe$^2$Net\cite{he2020fhde2net} & ECCV 2020  & 28.243 & 0.9654 & 0.0343 & 561.916 & 13.599 & 55.30  \\
    MZNet\cite{lee2025moir}          & arXiv 2025 & 27.504 & 0.9598 & 0.0408 & 19.374  & 7.276  & 21.52  \\
    P-BiC\cite{xiao2024p}            & ACM MM 2024& 28.590 & 0.9689 & 0.0334 & 17.304  & 4.922  & 38.17  \\
    ESDNet-L\cite{yu2022towards}     & ECCV 2022  & \underline{28.809} & \underline{0.9711} & \underline{0.0268} & 129.674 & 10.623 & 19.22  \\
    FaDeNet (Ours)                   & --         & \textbf{32.159} & \textbf{0.9859} & \textbf{0.0169} & 133.681 & 7.051  & 38.10  \\
    \bottomrule
  \end{tabular}
  }
\end{table}

\begin{figure}[t]
  \centering
  \includegraphics[width=\linewidth]{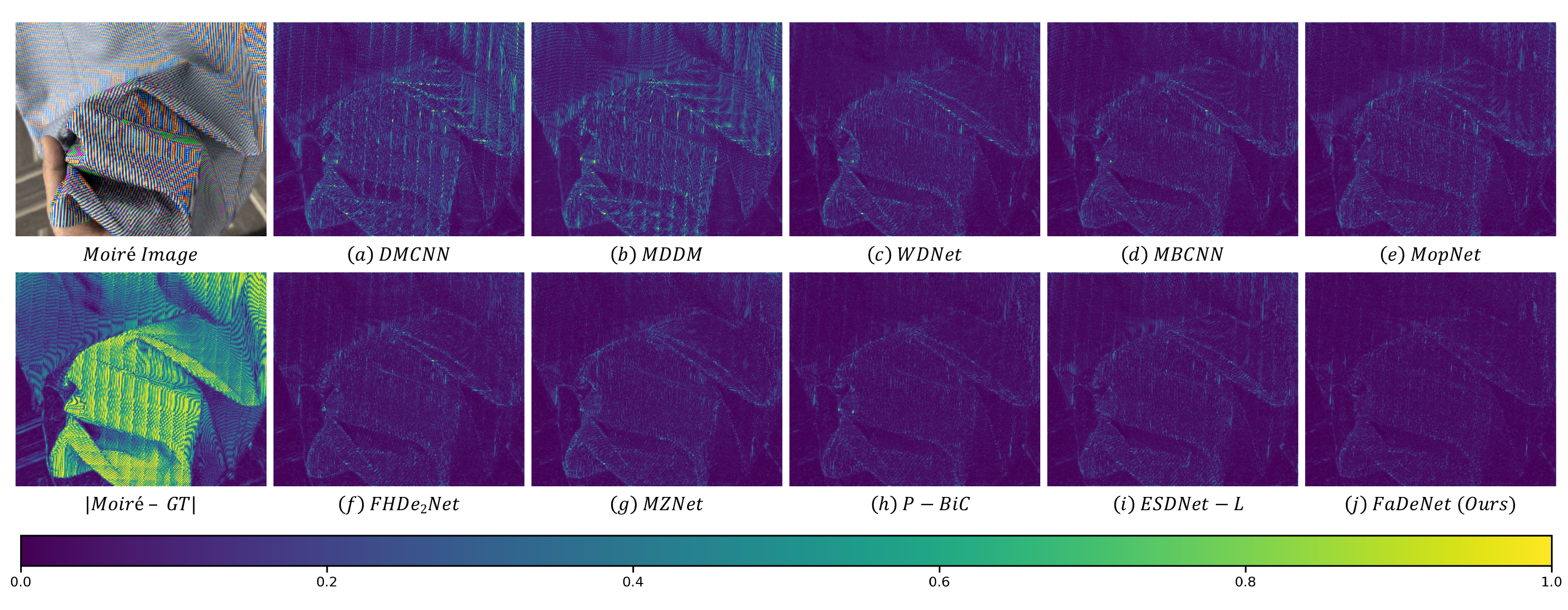}
   \caption{Visual comparison of error maps on the PRISM dataset. The color transition from dark purple to bright yellow represents increasing absolute errors ($| \text{Output} - \text{GT} |$).}
  \label{fig:prism_qual}
\end{figure}

\subsection{Comparison with Screen Demoir\'{e}ing Methods on PRISM}

\subsubsection{Quantitative results.}
\cref{tab:prism_quant} reports quantitative comparisons on PRISM. 
FaDeNet achieves the best performance across all three quality metrics. The large PSNR margin indicates that FaDeNet more effectively restores structural fidelity under strong moir\'{e} interference, while the lower LPIPS suggests better perceptual alignment, particularly important for fabric images with dense and repetitive textures.

\subsubsection{Qualitative results.}
\cref{fig:prism_qual} visualizes representative examples with absolute error maps. Compared with prior methods, FaDeNet produces the cleanest residual maps with fewer bright error regions, implying more complete removal of moir\'{e} components.  These visual results are consistent with the quantitative gains, confirming that FaDeNet improves structural accuracy and visual realism simultaneously.

\subsubsection{Computational cost analysis.}
Beyond restoration quality, FaDeNet maintains a competitive balance between architectural complexity and inference speed. As shown in Table \ref{tab:prism_quant}, FaDeNet is relatively compact with 7.05M parameters, fewer than the second-best performing model ESDNet-L (10.62M). In terms of throughput, FaDeNet achieves 38.10 FPS, nearly doubling the speed of ESDNet-L (19.22 FPS) despite having a similar GFLOPs count. These metrics suggest that FaDeNet provides a favorable trade-off in efficiency and quality.

\subsection{Generalization to Real-World Fabric Demoir\'{e}ing}
\begin{figure}[!t]
  \centering
  \includegraphics[width=\linewidth]{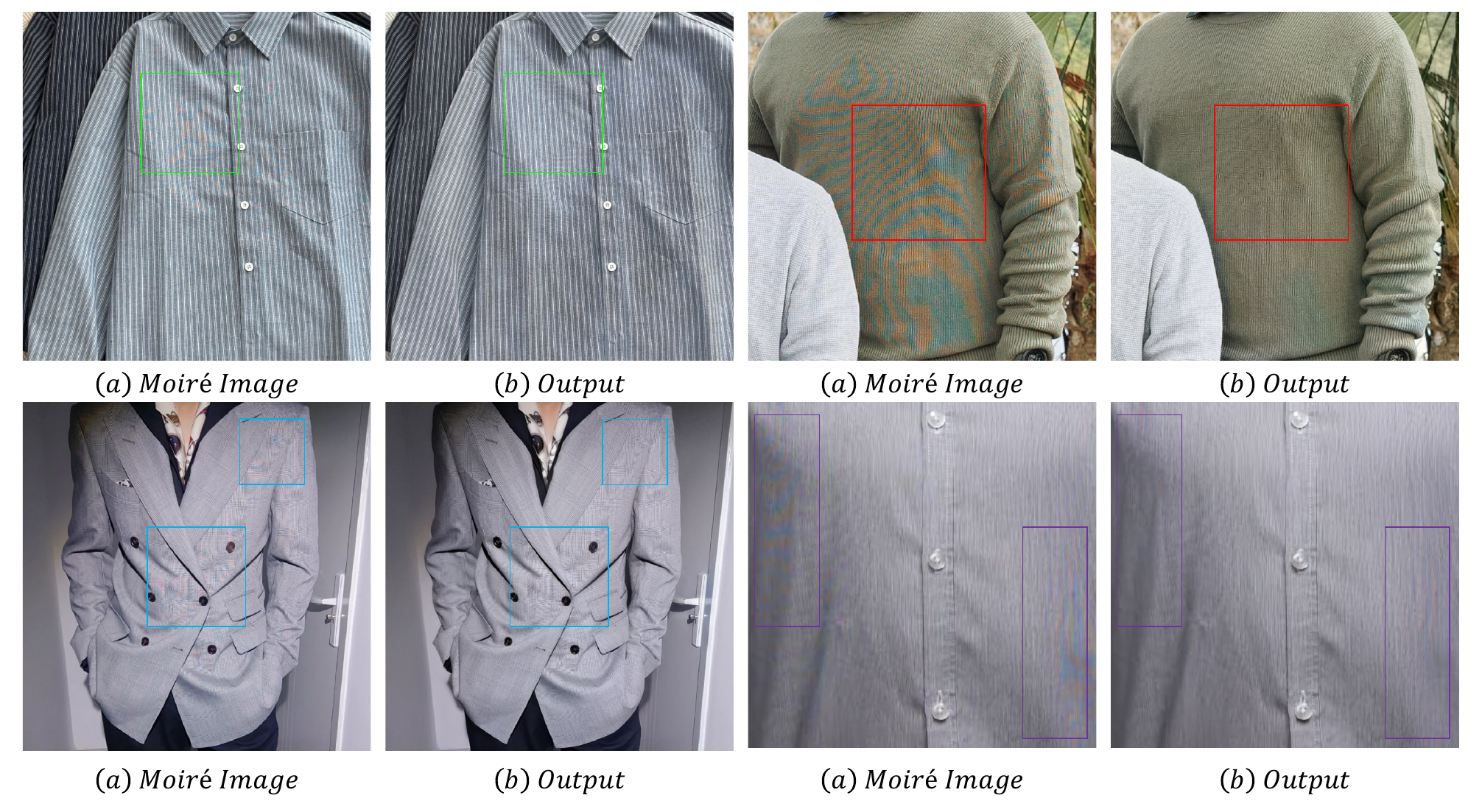}
  \caption{Qualitative results on real-world cases by models trained only on PRISM}
  \label{fig:real_generalization}
  \vspace{-9pt}
\end{figure}

To evaluate real-world robustness, we directly apply PRISM-trained models to an unpaired real-world fabric moir\'{e} dataset without any fine-tuning. As shown in Fig. \ref{fig:real_generalization}, PRISM-trained models produce visually meaningful results, effectively suppressing moir\'{e} artifacts while preserving plausible texture structures. To further quantify this zero-shot generalization, we report no-reference perceptual quality scores and human subjective results in Table \ref{tab:combined_nriqa_mos}.

\begin{table*}[!t]
\centering
\caption{No-reference perceptual quality comparison and blind MOS user study results on the real-world unpaired fabric moir\'{e} dataset. Bold and underlined values indicate the best and second-best results, respectively.}
\label{tab:combined_nriqa_mos}
\setlength{\tabcolsep}{5pt}
\renewcommand{\arraystretch}{1.1}
\resizebox{\linewidth}{!}{
\begin{tabular}{l cc c ccccc}
\toprule
\multirow{2}{*}{Method} & \multicolumn{2}{c}{No-Reference IQA} & & \multicolumn{5}{c}{User Study (MOS)} \\
\cmidrule(lr){2-3} \cmidrule(lr){5-9}
& ARNIQA $\uparrow$ & CLIP-IQA+ $\uparrow$ & & MOS $\uparrow$ & Std. & 95\% CI $\downarrow$ & Avg. Rank $\downarrow$ & Corr. $p$ \\
\midrule
DMCNN\cite{sun2018moire}         & 0.59850 & 0.43023 & & -- & -- & -- & -- & -- \\
MDDM\cite{cheng2019multi}        & 0.62414 & 0.42228 & & -- & -- & -- & -- & -- \\
WDNet\cite{liu2020wavelet}       & 0.63115 & 0.43618 & & -- & -- & -- & -- & -- \\
MBCNN\cite{zheng2020image}       & 0.62612 & 0.44005 & & -- & -- & -- & -- & -- \\
FHDe$^2$Net\cite{he2020fhde2net} & 0.62763 & 0.44991 & & -- & -- & -- & -- & -- \\
MopNet\cite{he2019mop}           & \underline{0.63358} & 0.45106 & & 2.86 & 1.21 & $\pm$0.14 & 3.30 & $2.64{\times}10^{-8}$ \\
MZNet\cite{lee2025moir}          & 0.62660 & 0.44923 & & 3.11 & 1.13 & $\pm$0.13 & \underline{2.63} & $8.25{\times}10^{-4}$ \\
P-BiC\cite{xiao2024p}            & 0.63327 & \textbf{0.48329} & & \underline{3.17} & 1.22 & $\pm$0.14 & 2.90 & $8.18{\times}10^{-3}$ \\
ESDNet-L\cite{yu2022towards}     & 0.63001 & 0.45139 & & 2.79 & 1.13 & $\pm$0.13 & 4.20 & $2.91{\times}10^{-10}$ \\
FaDeNet (Ours)                   & \textbf{0.63364} & \underline{0.47473} & & \textbf{3.39} & 1.07 & $\pm$0.12 & \textbf{1.97} & -- \\
\bottomrule
\end{tabular}
}
\end{table*}

\subsubsection{No-reference perceptual quality.}
FaDeNet achieves the highest ARNIQA score, indicating superior distortion suppression and improved perceptual quality on real-world inputs. It also attains a competitive CLIP-IQA+ score, suggesting that its restorations remain visually natural and semantically consistent.

\subsubsection{Blind MOS user study.}
To further assess subjective visual appeal, we conducted a blind Mean Opinion Score (MOS) study. We selected five representative models and evaluated them across 15 groups of representative real-world images. The study involved 20 participants, including 10 domain experts in image processing and 10 volunteers. As shown in Table \ref{tab:combined_nriqa_mos}, FaDeNet outperforms other methods in human preference, achieving the best MOS of 3.39 and a leading Average Rank of 1.97. This indicates that FaDeNet provides reliable restoration quality across diverse real-world scenarios.  To assess statistical significance, we further conduct paired Wilcoxon signed-rank tests between FaDeNet and each MOS baseline, followed by Holm-Bonferroni correction. The corrected $p$ values in Table~\ref{tab:combined_nriqa_mos} are all below 0.01, indicating statistically significant preference for FaDeNet.

\subsection{Ablation Studies}

\subsubsection{Base/Detail Decomposition.}
Configuration (A) removes the base/detail decomposition and the detail branch. The results in \cref{tab:ablation} indicate that separating low-frequency structure and high-frequency fabric details is beneficial.

\subsubsection{Mask Gating.} 
Configuration (B) disables mask gating by setting $\mathbf M \equiv1$. The performance degrades in (B), suggesting that mask gating and spatially selective updates are important for real fabric 
demoir\'{e}ing.

\subsubsection{SAGB Design.} Configurations (C–E) validate the SAGB’s architecture: replacing it with a standard ConvBlock or removing the spectral gate and spatial branch causes a consistent drop in performance. This confirms that the synergy between frequency-domain gating and directional spatial modeling is vital for addressing the complex, periodic nature of moiré patterns.

\subsubsection{Training Objective.} Configuration (F) employs $\ell_1$ loss only. The results in \cref{tab:ablation} show a performance drop, particularly in LPIPS, indicating that pixel-wise supervision alone is insufficient for high-frequency fidelity.

\begin{table}[t]
\centering
\caption{Ablation results of FaDeNet on PRISM.}
\label{tab:ablation}
\begin{tabular}{lccc}
\hline
{Configuration} & PSNR $\uparrow$ & SSIM $\uparrow$ & LPIPS $\downarrow$ \\
\hline
(A) w/o Decomposition              & 31.560 & 0.9837 & 0.0186 \\
(B) w/o Mask ($\mathbf M\equiv 1$)         & 31.599 & 0.9840 & 0.0185 \\
(C) w/o SAGB (param.-matched ConvBlock)                & 31.349 & 0.9831 & 0.0191 \\
(D) SAGB w/o spectral gate           & 31.541 & 0.9837 & 0.0189 \\
(E) SAGB w/o spatial anisotropic branch   & 31.355 & 0.9829 & 0.0194 \\
(F)  $\ell_1$ loss only                   & 31.777 & 0.9843 & 0.0183 \\
Full (FaDeNet)               & \textbf{32.159} & \textbf{0.9859} & \textbf{0.0169} \\
\hline
\end{tabular}
\end{table}

\section{Conclusion}

In this paper, we have systematically investigated the underexplored problem of fabric image demoiréing. We identify that the fundamental performance bottleneck of existing methods lies in the spectral entanglement between intrinsic textile micro-patterns and sampling-induced aliasing. To bridge this gap, we proposed PRISM, a physics-based residual injection pipeline that synthesizes realistic fabric moir\'{e} while guaranteeing perfect pixel-wise alignment and preserving native fabric details. We further introduced FaDeNet, a fabric demoir\'{e}ing architecture with base/detail decomposition and spectral-anisotropic gated blocks for jointly capturing spatial and frequency characteristics of moir\'{e}. Extensive experiments show that FaDeNet outperforms representative screen-camera demoir\'{e}ing baselines on PRISM and generalizes well to real unpaired fabric captures. We hope PRISM facilitates future research on fabric image demoir\'{e}ing.



\clearpage

%
%
\bibliographystyle{splncs04}
\bibliography{main}

\clearpage
\appendix

\setcounter{linenumber}{1}
\setcounter{page}{1}
\begin{center}
    {\LARGE\bfseries Supplementary Material}
\end{center}
\addcontentsline{toc}{section}{Supplementary Material}

\section{Distinguishing Screen Moir\'{e} from Fabric Moir\'{e}}

Although both screen moir\'{e} and fabric moiré are ultimately caused by sampling-induced aliasing, they should not be treated as the same restoration problem. As discussed in the main paper, the difference is not limited to visual appearance, but originates from the physical micro-structures being sampled and propagates through the image formation process to the final artifact distribution. In this section, we provide a comparison from three complementary perspectives: sampling source, sampling outcome, and frequency-domain behavior. Our goal is to clarify why screen-trained demoiréing models cannot be directly transferred to fabric scenes without suffering from residual artifacts or texture over-smoothing.

\subsection{Difference in Sampling Sources}
The first level distinction appears at the source level. Screen moiré originates from rigid planar displays with highly regular pixel or subpixel lattices. Such structures are globally stationary and strictly periodic, and their Fourier spectra often exhibit relatively sparse and discrete peaks. In contrast, fabric moiré arises from non-rigid textile surfaces with dense, anisotropic, and only semi-periodic weave microstructures. Real garments further introduce folds, stretching, local compression, and pose-dependent deformation, making the local sampling content spatially non-stationary even before aliasing occurs.

To visualize this difference, we compare representative source patches from the two domains together with their log-magnitude 2D Fourier spectra in \cref{fig:appendix1}. For fairness, all source patches are cropped at the same spatial size and analyzed with the same FFT normalization. The screen source exhibits a highly regular lattice pattern and a relatively sparse, grid-like spectrum with more localized peaks. By contrast, the fabric source presents more distributed spectral support, with broadened peaks, directional spread, and stronger patch-to-patch variation. This distinction is important because the restoration problem is already more ill-posed before the camera sampling stage: in fabrics, the intrinsic content itself occupies a broader and less separable frequency range.

\setcounter{figure}{0}
\begin{figure}[t]
  \centering
  \includegraphics[width=\linewidth]{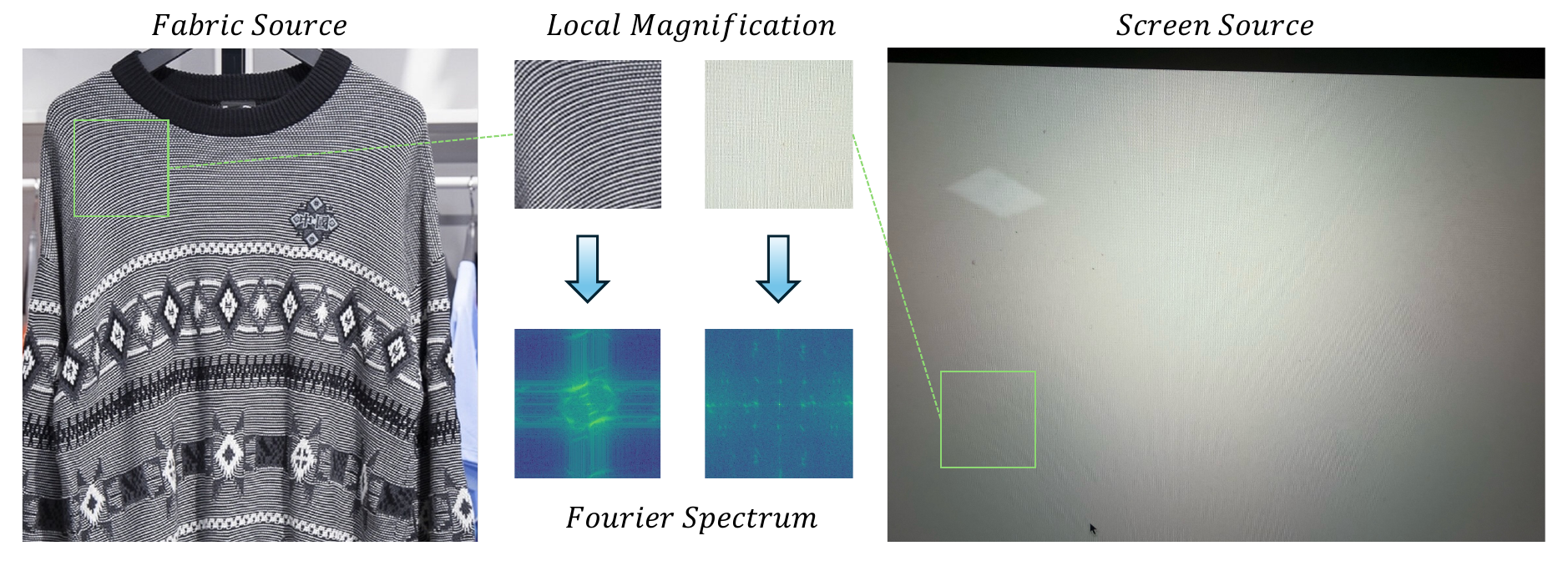}
  \caption{Representative source patches and corresponding log-magnitude 2D Fourier spectrum for screen and fabric domains. Under the same crop size and FFT normalization, the screen source shows a regular lattice structure and a relatively sparse, grid-like spectrum, whereas the fabric source exhibits broader and more distributed spectral support with stronger directional spread and local variation. This indicates that fabric content is intrinsically less stationary and less frequency-separable even before camera sampling.}
  \label{fig:appendix1}
  \vspace{-10pt}
\end{figure}

\subsection{Difference in Sampling Outcomes}

The second distinction appears in the observed aliasing patterns after image formation. Screen moiré is typically more global and more content-agnostic: once the display lattice and the camera sampling lattice become mismatched, the interference pattern tends to remain stable over large image regions and is not tightly tied to semantic image content. By contrast, fabric moiré is usually more localized, emerging mainly in highly textured textile regions and being strongly influenced by local weave geometry, deformation, and shading.

\begin{figure}[t]
  \centering
  \includegraphics[width=\linewidth]{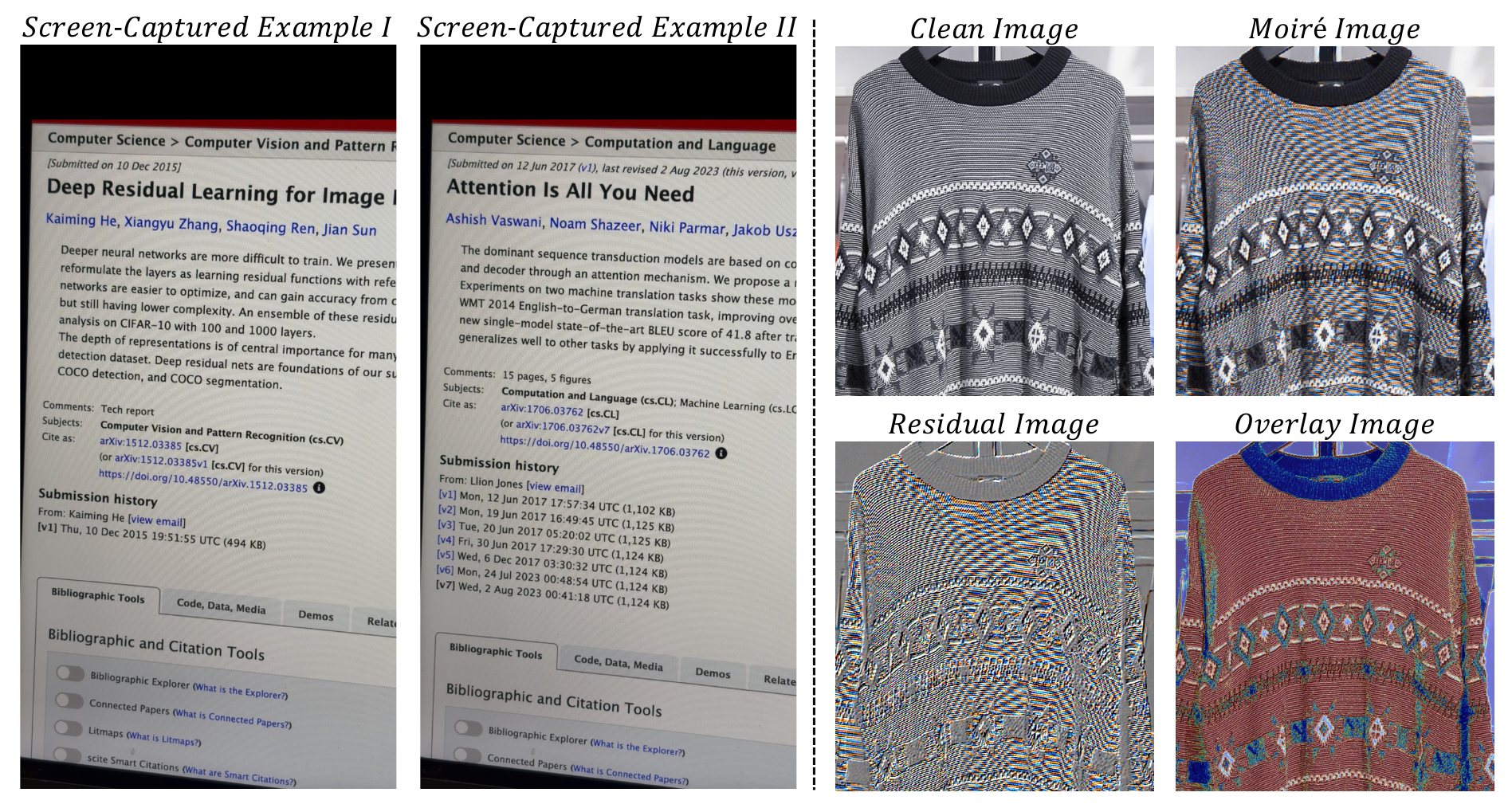}
  \caption{Comparison of sampling outcomes in screen and fabric moiré. Left: screen moiré remains largely unchanged when the displayed content is replaced, indicating that the artifact is mainly governed by the mismatch between the display lattice and the camera sampling lattice, rather than by semantic image content. Right: for fabric moiré, we visualize the clean image, the moiré image, the estimated residual, and the residual overlay. The artifact is concentrated in textile-rich regions and follows local fabric structure, suggesting that fabric moiré is more localized and more tightly coupled with the underlying texture field.}
  \label{fig:appendix2}
\end{figure}

To illustrate this difference, we conduct two qualitative comparisons in \cref{fig:appendix2}.. For the screen case, we capture screen moiré while changing the displayed image content, and observe that the overall moiré pattern remains largely unchanged. This suggests that screen moiré is dominated by the interaction between the display lattice and the camera sampling lattice, rather than by the semantic content shown on the screen. For the fabric case, we compare the clean image, the moir\'{e} observation, the estimated residual, and the residual overlay. 
Unlike screen moir\'{e}, the residual is concentrated in textile-rich regions and tends to follow local fabric structures such as weave orientation, folds, and deformation.

These observations suggest that fabric moiré is not well described as a simple content-independent foreground layer. Instead, it should be understood as an aliasing artifact entangled with the intrinsic textile texture, which makes restoration more challenging than in the screen setting.

\subsection{Difference in Frequency-Domain Behavior}

The third distinction lies in the frequency domain. Although both screen and fabric moir\'{e} originate from sampling-induced aliasing, their spectral organization can still differ noticeably in practice. In particular, fabric moir\'{e} is formed on top of dense and semi-periodic textile structures, which often leads to a broader and less concentrated frequency distribution.

To provide a simple quantitative observation, we compute normalized spectral entropy under a unified protocol. Specifically, each moir\'{e} image is converted to grayscale, center-cropped to \(128\times128\), mean-subtracted, multiplied by a 2D Hann window, and then transformed by 2D FFT. Let \(A(u,v)\) denote the magnitude spectrum and let
\[
\hat{A}(u,v)=\frac{A(u,v)}{\sum_{u,v}A(u,v)}
\]
be its normalized form. The spectral entropy is defined as
\begin{equation}
\mathcal{H}=-\sum_{u,v}\hat{A}(u,v)\log \hat{A}(u,v).
\end{equation}
Higher entropy indicates that the frequency energy is distributed over a broader support, while lower entropy corresponds to a more concentrated and structured spectrum.

As shown in \cref{tab:spectral_entropy}, PRISM obtains a mean normalized spectral entropy of 0.910, which is substantially higher than FHDMi (0.838) and TIP2018 (0.834). This suggests that the moir\'{e} patterns in PRISM occupy a broader and more distributed frequency regime, rather than being dominated by relatively sparse and regular spectral peaks. Such an observation is consistent with the intuition that fabric moir\'{e} is more tightly entangled with complex textile textures and therefore exhibits higher frequency-domain complexity.

We note that spectral entropy is only used here as an auxiliary statistic in the appendix, rather than a complete characterization of the fabric-screen domain gap. The distinction between the two tasks should still be understood jointly from the sampling source, the spatial manifestation of the artifacts, and their frequency-domain behavior.

\setcounter{table}{0}
\begin{table}[t]
  \centering
  \caption{Normalized spectral entropy on different moir\'{e} datasets under a unified protocol. Each image is converted to grayscale, center-cropped to $128\times128$, mean-subtracted, multiplied by a 2D Hann window, and analyzed using the FFT magnitude spectrum. Higher values indicate a broader and more distributed frequency support.}
  \label{tab:spectral_entropy}
  \setlength{\tabcolsep}{10pt}
  \renewcommand{\arraystretch}{1.1}
  \begin{tabular}{lcc}
    \toprule
    Dataset & Mean $\uparrow$ & Std. \\
    \midrule
    PRISM    & \textbf{0.910} & 0.040 \\
    FHDMi    & 0.838 & 0.066 \\
    TIP2018  & 0.834 & 0.051 \\
    \bottomrule
  \end{tabular}
\end{table}

\section{Additional Details of PRISM}

\subsection{GT Data Collection}
To construct a clean GT pool for fabric image demoir\'{e}ing, we first collected 42,939 candidate garment images from the Internet. Since our goal is to focus on clothing patterns that are highly prone to moir\'{e} under real-world camera sampling, the search process was guided by fabric categories with dense and repetitive structures. Specifically, the keywords included striped shirts, checked shirts, striped suits, checked suits, houndstooth sweaters, houndstooth shirts, fine horizontal striped knitwear, dense jacquard knitwear, fine checked knit tops, pixel-pattern jacquard sweaters, and densely woven geometric jacquard knitwear.

We then applied a multi-stage data cleaning pipeline to improve both content relevance and image quality. First, duplicate images were removed using a combination of exact MD5 matching and perceptual hash (PHash) matching. We further discarded images that were excessively small or had unreasonable aspect ratios. Next, we localized and cropped the clothing regions so that the retained samples focused on the garments themselves. To reduce textual contamination, we identified and removed images containing text. Finally, we performed semantic filtering followed by manual verification, retaining only visually clean garment images with clear textile structures and no obvious non-fabric interference.

To further ensure that the retained GT images are visually moir\'{e}-free, all candidate clean images were independently inspected by two domain experts. We additionally performed an FFT-based screening step to flag images with suspicious periodic aliasing patterns for re-inspection. This screening step flagged 40 images, none of which was confirmed as moir\'{e}-contaminated after expert re-inspection. After this cleaning and verification process, 1,963 clean images were retained as the GT pool, including 1,783 images for training and 180 images for testing.

\subsection{Pair Synthesis}
\begin{figure}[t]
\centering
\includegraphics[width=\linewidth]{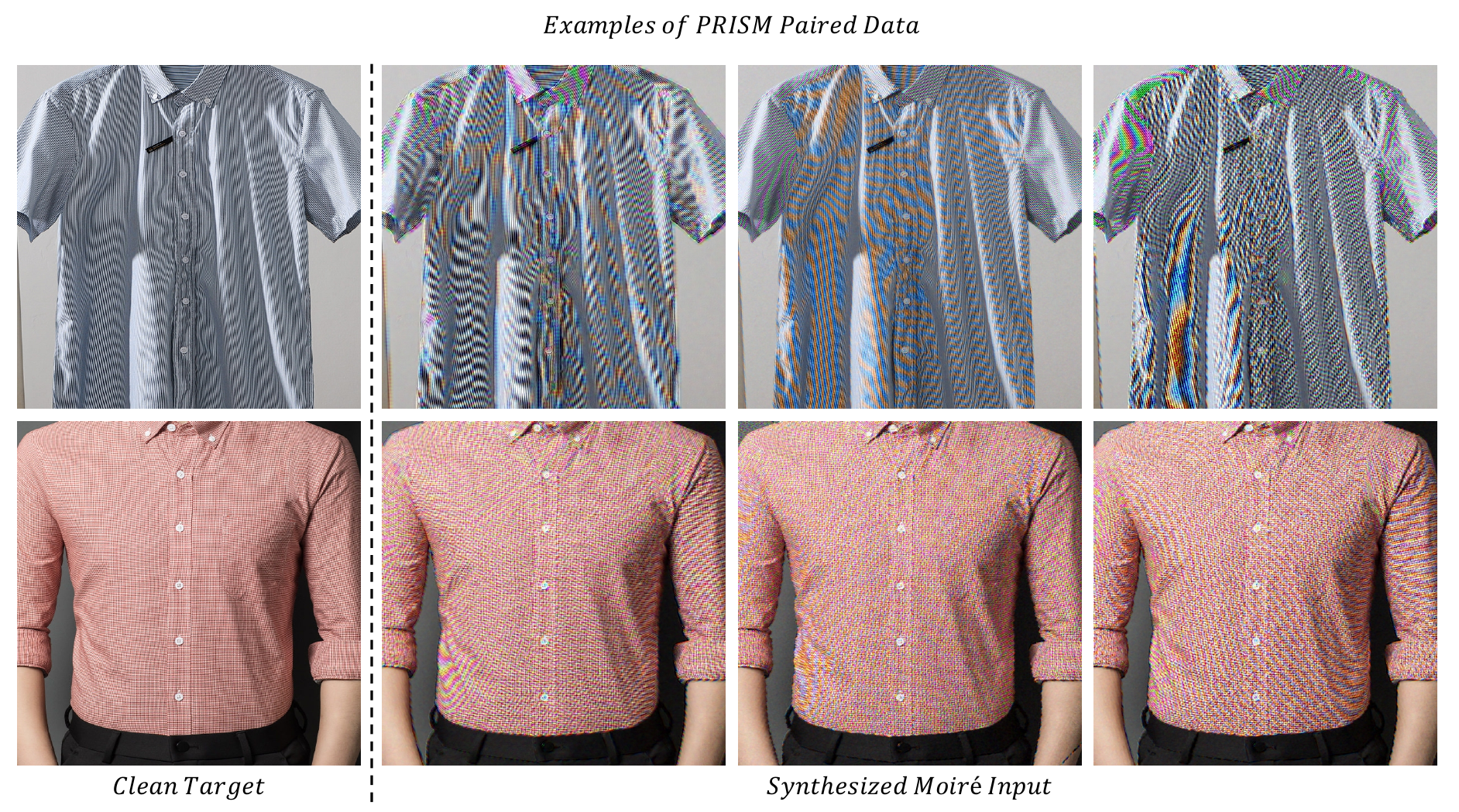}
\caption{PRISM paired data examples. Each clean GT image is paired with multiple synthesized moir\'{e} observations under different sampled imaging parameters.}
\label{fig:prism_pair_examples}
\end{figure}

Based on these GT images, we further constructed paired data using the proposed PRISM pipeline. For each GT image, we adopted a one-to-many synthesis strategy and generated up to 15 moir\'{e} variants under different randomly sampled physical parameters. Specifically, the synthesis process randomly perturbs geometric scale, in-plane rotation, and residual injection strength, while also introducing stochastic variations in the sampling and ISP simulation stages. This design allows each GT image to produce multiple moir\'{e} observations with diverse aliasing patterns and artifact severities, better reflecting the variability of real fabric capture conditions. Representative PRISM paired examples are shown in \cref{fig:prism_pair_examples}.

To improve the quality and usefulness of the synthesized pairs, we further filtered the generated samples according to their PSNR values with respect to the corresponding GT images. In particular, we retained only those synthesized images whose PSNR lies in the range of $[10, 20]$, which excludes cases with overly weak degradation as well as cases with excessively severe degradation. In addition, to maintain sufficient diversity for each clean image, we enforced that each GT image contributes at least three valid moir\'{e} inputs after filtering. Following this procedure, we obtained 14,587 training pairs and 1,463 testing pairs, resulting in 16,050 paired samples in total.

This pair synthesis strategy provides a controllable yet diverse benchmark for fabric demoir\'{e}ing, where each clean garment image is associated with multiple moir\'{e} observations of different severity levels and spectral characteristics.

\section{Additional Details of FaDeNet}
This section provides additional implementation details of FaDeNet from three aspects: the content-adaptive base/detail decomposition module, the Spectral-Anisotropic Gated Block (SAGB), and the scale-wise parameter allocation. Unless otherwise specified, parameter counts include learnable weights and biases only. Non-learnable operations, such as FFT, padding, element-wise combinations, and activation functions, are counted as zero. For depthwise convolutions, only kernel weights are counted, and for GroupNorm, two affine parameters are counted per channel.

\subsection{Content-Adaptive Base/Detail Decomposition}

The content-adaptive base/detail decomposition module uses a learnable bias-free depthwise Gaussian blur (kernel size $5$, $\sigma=1.2$) together with a lightweight gating subnetwork. Concretely, the gate is implemented as a two-layer convolutional stack composed of a $3\times3$ convolution ($3\rightarrow8$), a GELU activation, and a second $3\times3$ convolution ($8\rightarrow1$), followed by a sigmoid function to produce the single-channel gating map $\mathbf{W}$. The exact layer configuration and parameter counts are listed in \cref{tab:bd_layers}.

\begin{table}[t]
\centering
\caption{Layer configuration of the content-adaptive base/detail decomposition module for RGB input.}
\label{tab:bd_layers}
\setlength{\tabcolsep}{6pt}
\renewcommand{\arraystretch}{1.1}
\begin{tabular}{lcccc}
\toprule
Layer & In/Out Ch. & Kernel & Stride & Params \\
\midrule
Gate Conv1 & $3 \rightarrow 8$ & $3\times3$ & 1 & 224 \\
GELU & $8 \rightarrow 8$ & -- & -- & 0 \\
Gate Conv2 & $8 \rightarrow 1$ & $3\times3$ & 1 & 73 \\
Sigmoid & $1 \rightarrow 1$ & -- & -- & 0 \\
Depthwise Gaussian Conv (groups=3) & $3 \rightarrow 3$ & $5\times5$ & 1 & 75 \\
\midrule
Total & -- & -- & -- & 372 \\
\bottomrule
\end{tabular}
\end{table}

As shown in \cref{tab:bd_layers}, this module is highly lightweight, introducing only 372 learnable parameters for RGB input. Most of the capacity lies in the gating subnetwork, while the depthwise Gaussian filtering itself contributes only a small number of parameters. This is consistent with our design goal of explicitly separating low-frequency structure from high-frequency textile detail with negligible computational overhead.

\subsection{Spectral-Anisotropic Gated Block}
The Spectral-Anisotropic Gated Block (SAGB) first applies GroupNorm and a $1\times1$ channel expansion to $4C$, followed by four depthwise branches for directional and multi-scale spatial filtering. These are then fused by a $1\times1$ projection layer. In parallel, the frequency branch uses a lightweight FFT-based channel gate with hidden width $H=\max(8,C/4)$. Finally, the spatial and frequency features are concatenated and used to predict a gating map and a residual update. The layer-wise parameterization of a single SAGB is summarized in \cref{tab:sagb_layers}.

\begin{table}[t]
\centering
\caption{Layer configuration of the Spectral-Anisotropic Gated Block (SAGB), where $C$ denotes the block width and $H=\max(8,C/4)$.}
\label{tab:sagb_layers}
\setlength{\tabcolsep}{6pt}
\renewcommand{\arraystretch}{1.1}
\begin{tabular}{lcccc}
\toprule
Layer & In/Out Ch. & Kernel & Stride & Params \\
\midrule
GroupNorm (affine) & $C \rightarrow C$ & -- & -- & $2C$ \\
Expand Conv & $C \rightarrow 4C$ & $1\times1$ & 1 & $4C^2+4C$ \\
DW Conv & $C \rightarrow C$ & $3\times3$ & 1 & $9C$ \\
DW Conv & $C \rightarrow C$ & $1\times5$ & 1 & $5C$ \\
DW Conv & $C \rightarrow C$ & $5\times1$ & 1 & $5C$ \\
DW Conv (d=2) & $C \rightarrow C$ & $3\times3$ & 1 & $9C$ \\
Project Conv & $4C \rightarrow C$ & $1\times1$ & 1 & $4C^2+C$ \\
FFT Gate Conv & $C \rightarrow H$ & $1\times1$ & 1 & $CH+H$ \\
FFT Gate Conv & $H \rightarrow C$ & $1\times1$ & 1 & $CH+C$ \\
Gate Conv & $2C \rightarrow C$ & $1\times1$ & 1 & $2C^2+C$ \\
Update Conv & $2C \rightarrow C$ & $1\times1$ & 1 & $2C^2+C$ \\
\bottomrule
\end{tabular}
\end{table}

From \cref{tab:sagb_layers}, it can be seen that the main learnable capacity of SAGB comes from the $1\times1$ channel mixing layers, while the directional spatial operators remain parameter-efficient due to their depthwise design. This matches the role of SAGB in FaDeNet: to combine anisotropic spatial modeling with selective frequency-aware modulation without introducing excessive overhead.

\subsection{Scale-wise Parameter Allocation}
FaDeNet uses four feature scales with channel widths $C=\{48,96,192,384\}$. Following the design in the main paper, the FFT-based spectral gate is enabled only at the coarser $1/4$ and $1/8$ scales. The corresponding per-block parameter counts are listed in \cref{tab:sagb_params}.

\begin{table}[t]
\centering
\caption{Per-block parameter counts of SAGB at different scales in FaDeNet.}
\label{tab:sagb_params}
\setlength{\tabcolsep}{8pt}
\renewcommand{\arraystretch}{1.1}
\begin{tabular}{ccccc}
\toprule
Scale & $C$ & $H$ & FFT gate & Params / block \\
\midrule
$1$   & 48  & 12 & No  & 29,424 \\
$1/2$ & 96  & 24 & No  & 114,144 \\
$1/4$ & 192 & 48 & Yes & 468,144 \\
$1/8$ & 384 & 96 & Yes & 1,857,888 \\
\bottomrule
\end{tabular}
\end{table}

As the feature width increases, the parameter count of each SAGB grows accordingly, with the majority of the model capacity concentrated at the coarser scales. This allocation is intentional: coarse-resolution stages provide larger effective receptive fields and are better suited for capturing structured moir\'{e} interference and frequency-domain regularities, while fine-resolution stages focus more on conservative texture-preserving refinement.

\section{Additional Experimental Results}
\subsection{Mask Visualization and Analysis}

\begin{figure*}[t]
    \centering
    \includegraphics[width=\linewidth]{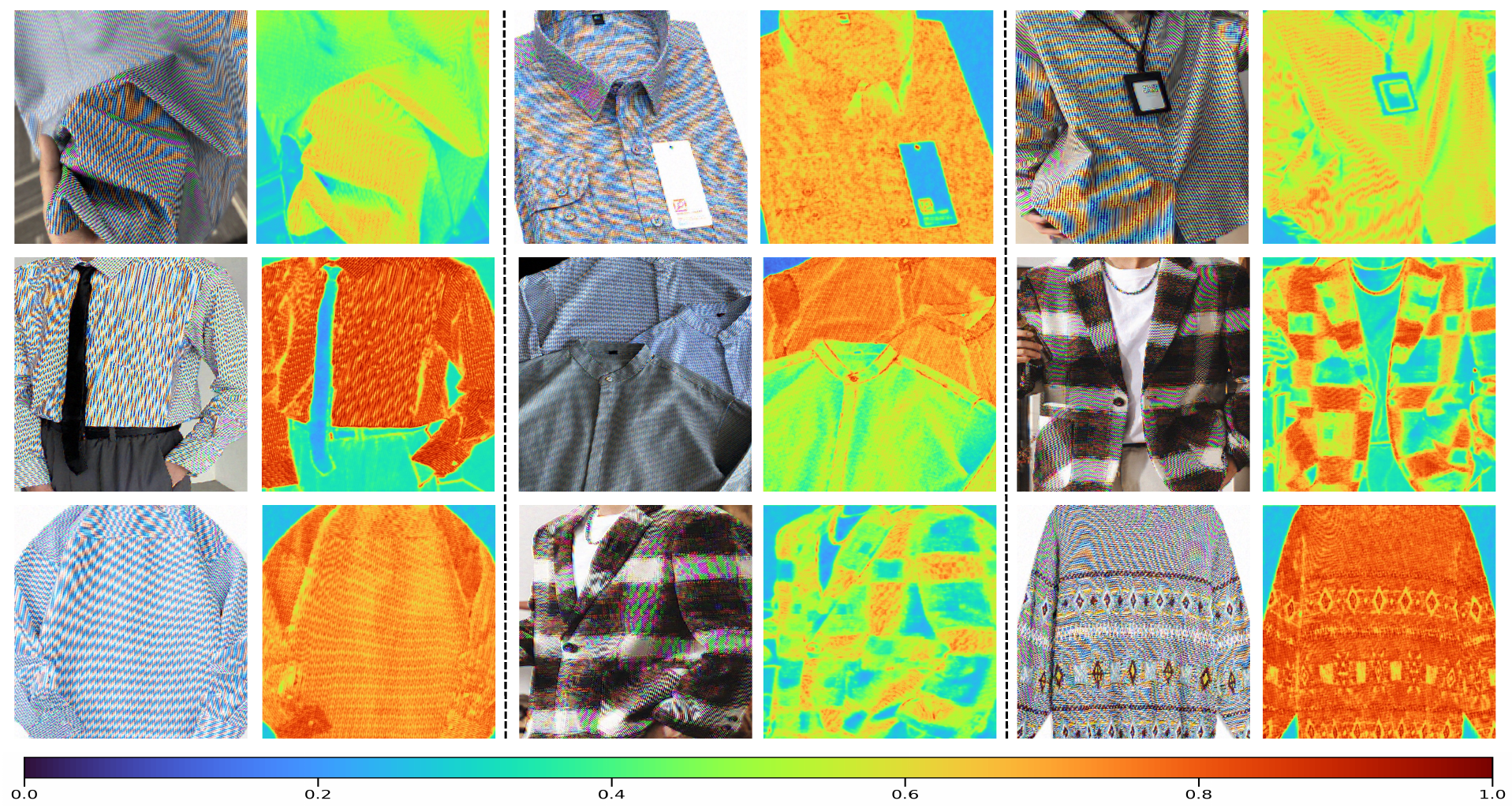}
    \caption{Visualization of learned spatial confidence masks on representative PRISM training samples. For each example, the left image is the moir\'{e} input and the right image is the corresponding predicted mask. Warmer colors indicate higher confidence and therefore stronger correction strength. The learned masks mainly focus on textile-rich regions with dense repetitive structures and strong moir\'{e} contamination, while assigning relatively weak responses to clean regions and non-fabric areas. This confirms that the mask-gated design in FaDeNet performs spatially selective restoration rather than applying uniform corrections over the entire image.}
    \label{fig:mask_vis}
    \vspace{-10pt}
\end{figure*}

To better understand the behavior of the proposed mask-gated restoration mechanism, we visualize the learned spatial confidence masks predicted by FaDeNet on representative samples from the PRISM training set in \cref{fig:mask_vis}. For each example, we show the moir\'{e} input together with its corresponding mask response, where warmer colors indicate higher confidence and thus stronger correction strength.

Several observations can be made. First, the predicted masks are not uniformly activated over the entire image, but instead concentrate on fabric regions that exhibit strong moir\'{e} contamination. In particular, highly textured garments with dense stripes, checks, or repetitive woven patterns tend to receive consistently higher responses, indicating that the network learns to identify regions where structured aliasing is more likely to dominate. By contrast, smooth background areas and non-fabric objects usually receive weaker responses, showing that the mask suppresses unnecessary modifications in visually clean regions.

Second, the mask is also sensitive to the spatial distribution and severity of the artifacts. For relatively mild cases, the response is more localized and sparse, whereas for images with globally strong interference the mask becomes broader and more active over most of the garment area. This behavior is consistent with the design goal of FaDeNet: instead of applying uniform restoration everywhere, the model selectively allocates stronger corrections to moir\'{e}-dominant regions while preserving clean textures as much as possible.

Third, the mask does not simply follow object silhouettes. In several examples, different regions within the same garment exhibit clearly different response levels, suggesting that the learned gating mechanism is driven not only by semantic garment location but also by local texture frequency, moir\'{e} strength, and structural complexity. This supports our claim that the mask acts as a content-adaptive controller for conservative restoration, helping the network balance artifact suppression against texture fidelity.

\vspace{-10pt}

\subsection{Failure Cases and Limitations}
\begin{figure}[t]
\centering
\includegraphics[width=0.7\linewidth]{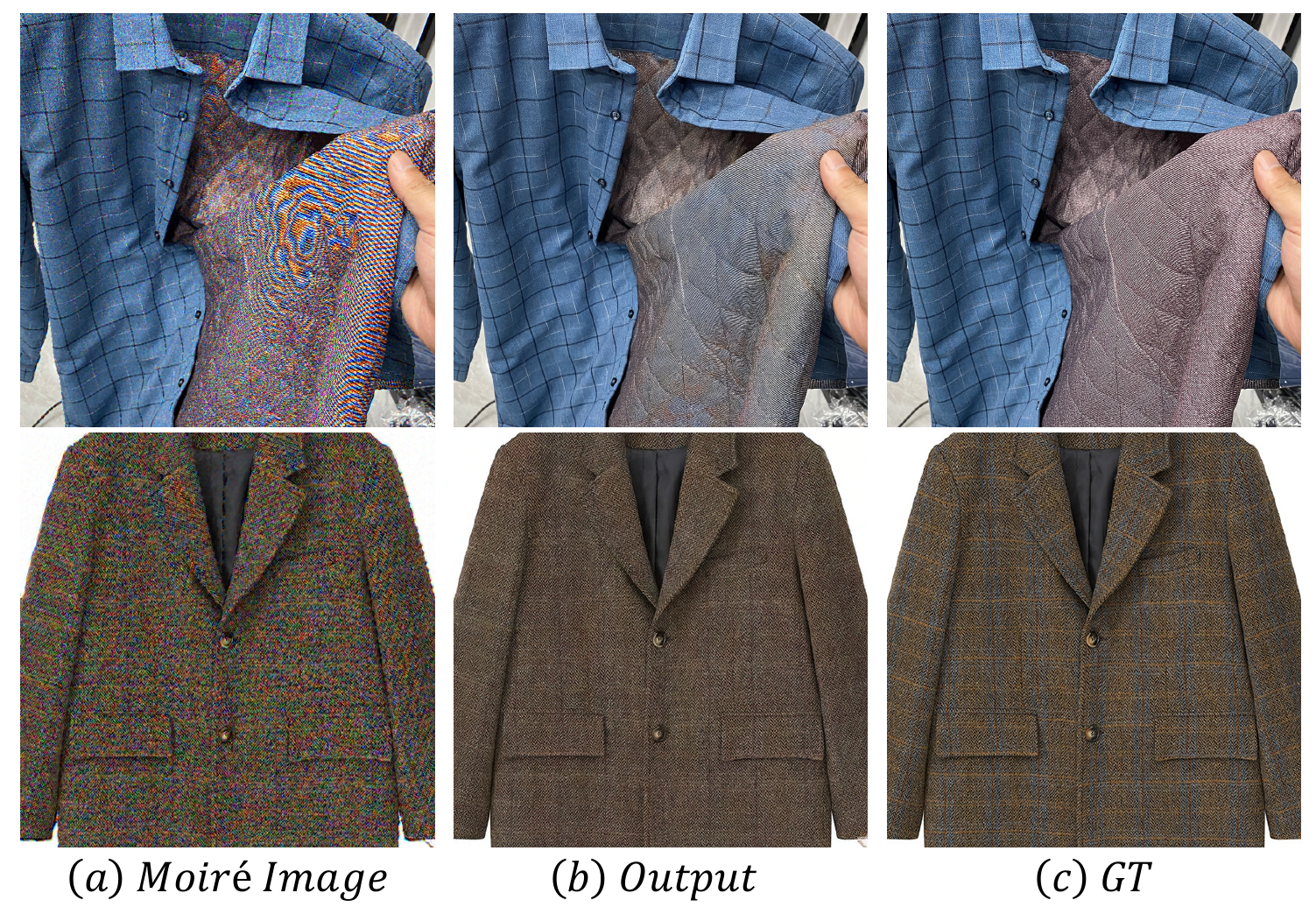}
\caption{Failure cases.}
\label{failure_case}
\vspace{-14pt}
\end{figure}

Although FaDeNet achieves good performance on both PRISM and real-world fabric moir\'{e} images, several challenging cases remain. As shown in \cref{failure_case}, fine repetitive textile structures may still lead to local restoration errors when the moir\'{e} pattern is highly entangled with the underlying fabric texture. In addition, when colored moir\'{e} artifacts overlap with authentic garment patterns, the model may over-suppress textile details while removing the artifact. These cases suggest that fabric demoir\'{e}ing remains challenging under severe spectral entanglement, and future work may further explore stronger texture-aware priors and real-world adaptation strategies.

\subsection{More Visual Results on PRISM}

\begin{figure}[t]
    \centering
    \includegraphics[width=\linewidth]{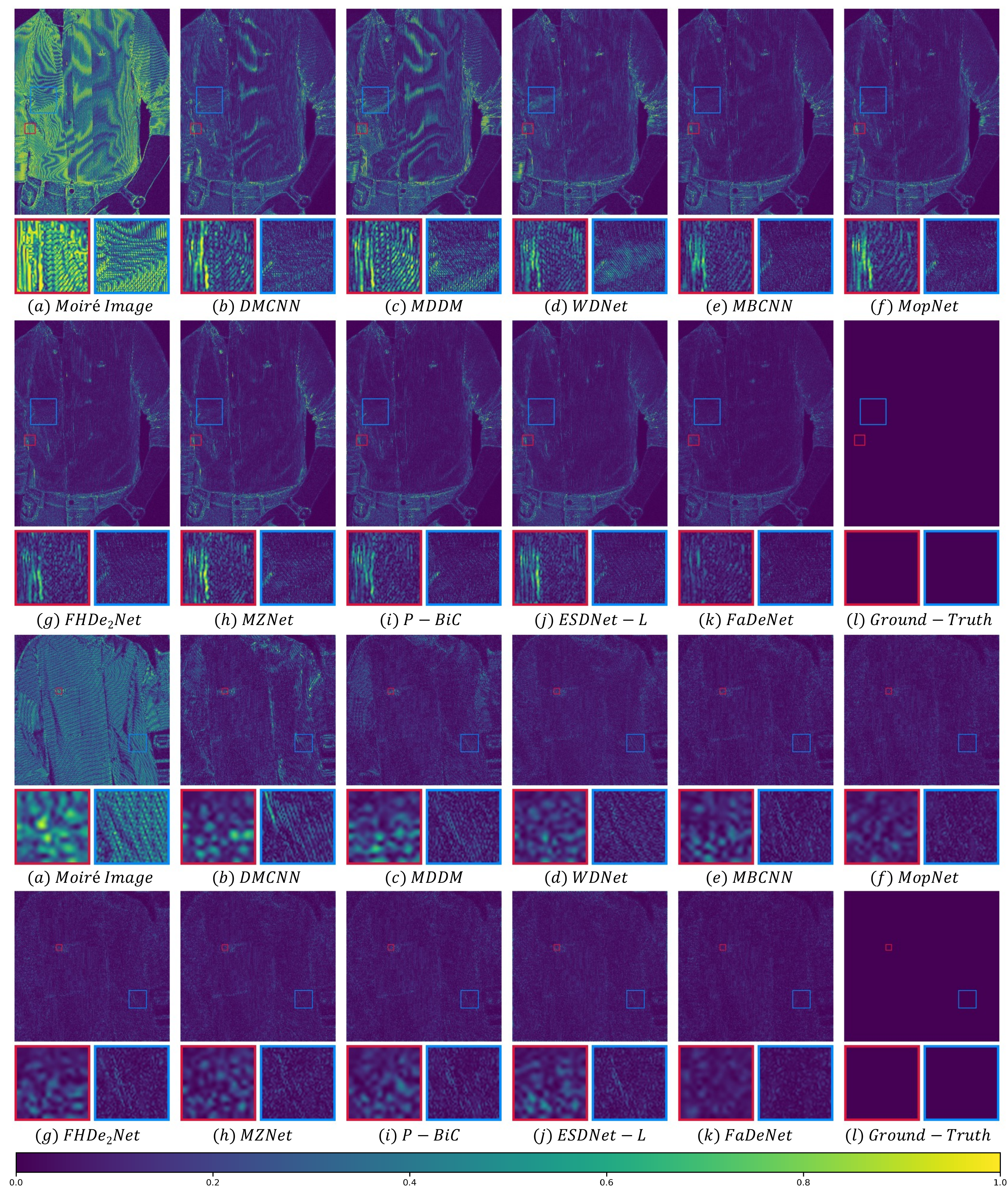}
    \caption{Visual comparison of error maps on the PRISM dataset (Part I).}
    \label{fig:app_prism_vis1}
\end{figure}

\begin{figure}[t]
    \centering
    \includegraphics[width=\linewidth]{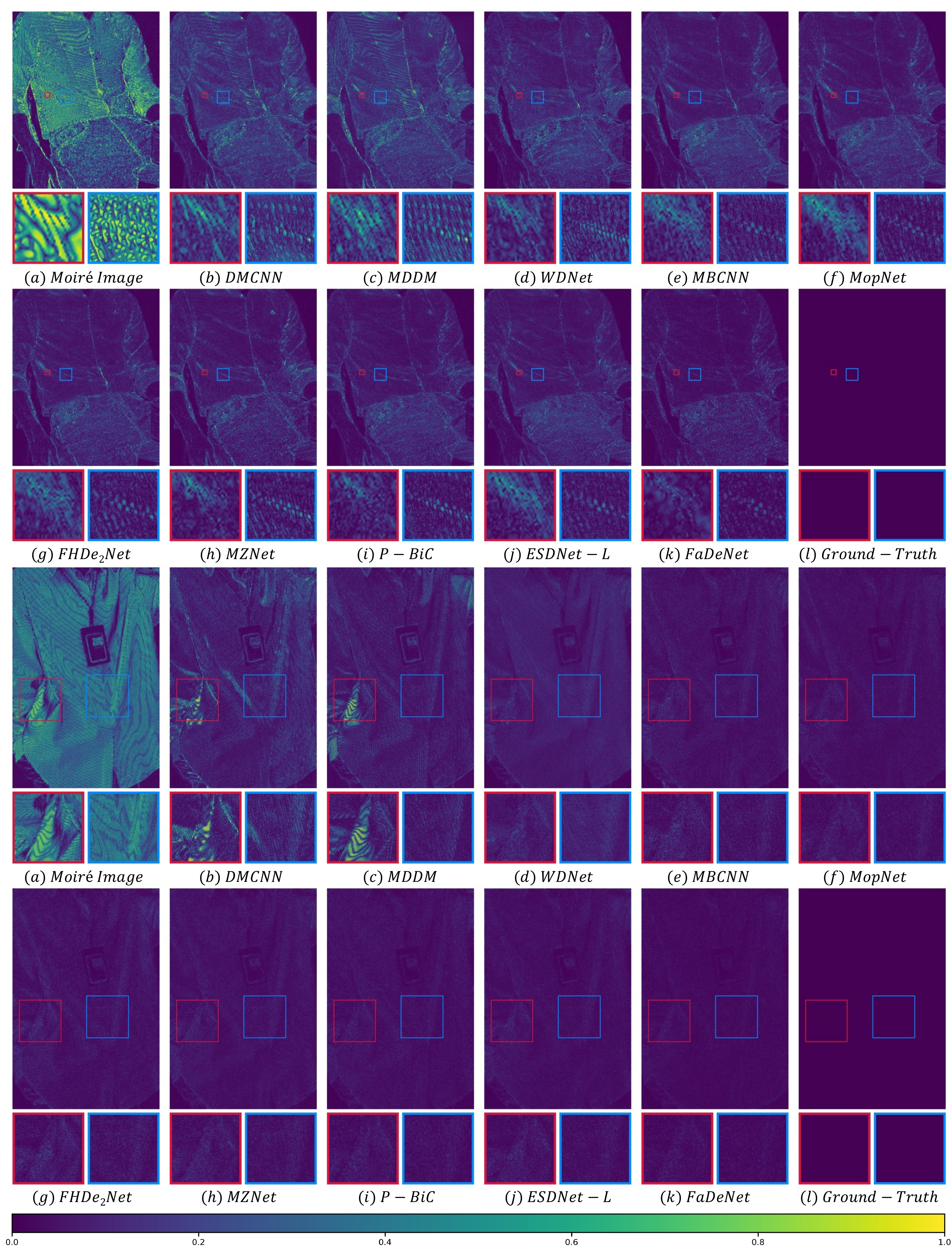}
    \caption{Visual comparison of error maps on the PRISM dataset (Part II).}
    \label{fig:app_prism_vis2}
\end{figure}

\begin{figure}[t]
    \centering
    \includegraphics[width=\linewidth]{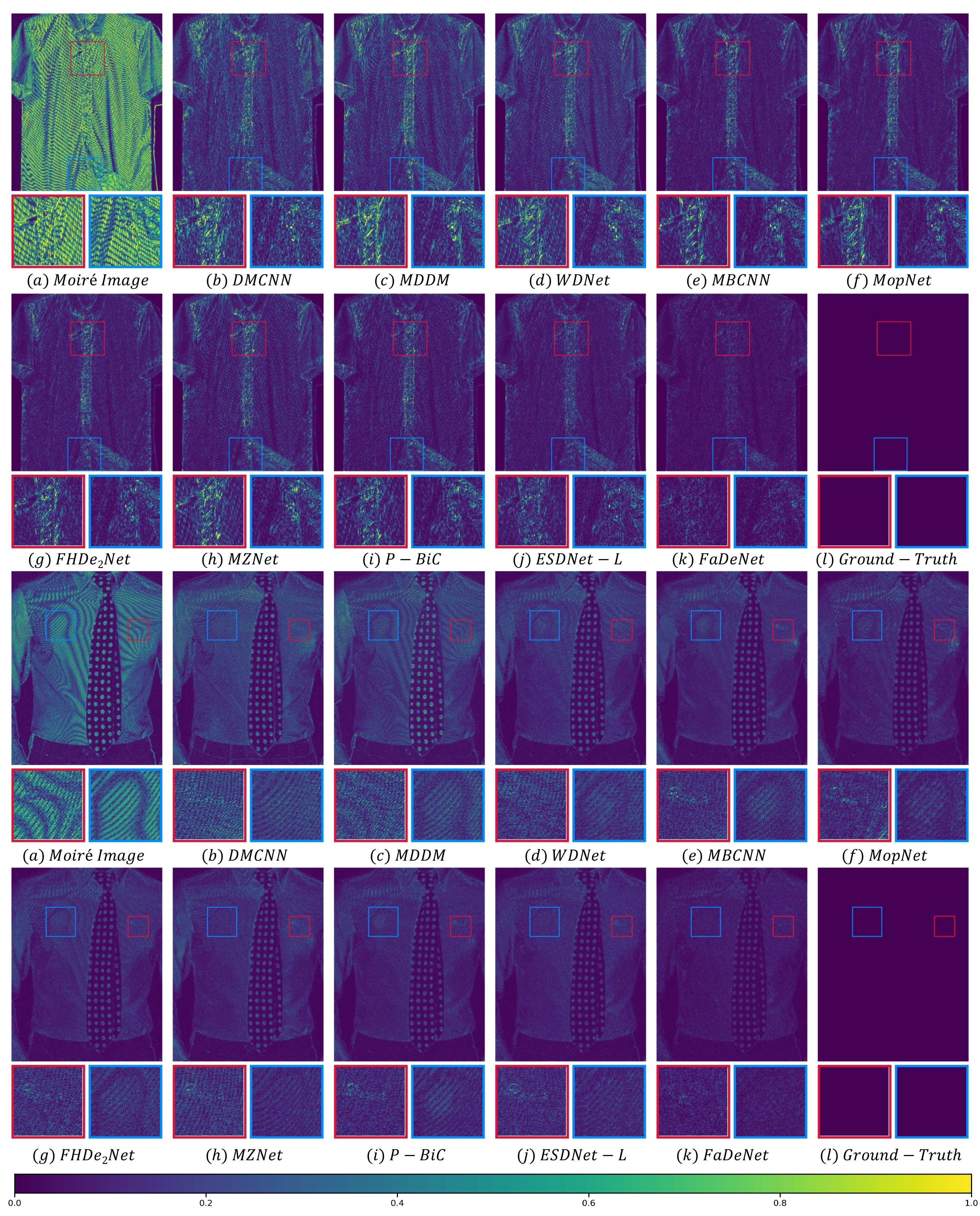}
    \caption{Visual comparison of error maps on the PRISM dataset (Part III)}
    \label{fig:app_prism_vis3}
\end{figure}

To provide a more comprehensive qualitative comparison, we present additional restoration results on the PRISM dataset in \cref{fig:app_prism_vis1,fig:app_prism_vis2,fig:app_prism_vis3}.

\end{document}